\documentclass{article}

\PassOptionsToPackage{numbers, compress}{natbib}

    \usepackage[preprint]{neurips_2019}

\usepackage[utf8]{inputenc} 
\usepackage[T1]{fontenc}    
\usepackage{hyperref}       
\usepackage{url}            
\usepackage{booktabs}       
\usepackage{amsfonts}       
\usepackage{nicefrac}       
\usepackage{microtype}      
\usepackage{graphicx}
\usepackage{subfig}
\usepackage{booktabs} 
\usepackage{ amssymb }
\usepackage{amsthm}
\usepackage{amsmath}
\usepackage{arydshln}
\usepackage{xcolor}
\usepackage[utf8]{inputenc}
\usepackage{url}
\usepackage[demo]{rotating}
\usepackage{amssymb}
\usepackage{rotating}

\newcommand{\vctr}[1]{\ensuremath{\pmb{#1}}}
\newcommand{\mtrx}[1]{\ensuremath{\pmb{#1}}}
\newcommand{\DeltaT}{\Delta t}
\newcommand{\timetovec}{Time2Vec}
\newcommand{\ttov}{\text{\textbf{t2v}}}
\newcommand{\periodicF}{\ensuremath{\mathcal{F}}}
\newcommand{\ie}{\emph{i.e.~}}
\newcommand{\eg}{\emph{e.g.}}

\newcommand{\ntdigits}{N\_TIDIGITS18}
\newcommand{\recallten}{Recall@10}
\newcommand{\mrrten}{MRR@10}
\newcommand{\lastfm}{Last.FM}
\newcommand{\culike}{CiteULike}
\newcommand{\emnist}{Event-MNIST}
\newcommand{\sine}{sine}
\newcommand{\sigmoid}{Sigmoid}
\newcommand{\tanhyp}{Tanh}
\newcommand{\relu}{ReLU}

\newtheorem{proposition}{Proposition}
\newtheorem{defn}{Definition}

\title{Time2Vec: Learning a Vector Representation of Time}

\author{%
  Seyed Mehran Kazemi\thanks{Equal contribution.},~~Rishab Goel$^*$,~~Sepehr Eghbali$^*$,~~Janahan Ramanan,~~Jaspreet Sahota, \\ \textbf{Sanjay Thakur}, \textbf{Stella Wu},~~\textbf{Cathal Smyth},~~\textbf{Pascal Poupart},~~\textbf{Marcus Brubaker} \\
  Borealis AI \\
}

\begin{document}

\maketitle

\begin{abstract}
Time is an important feature in many applications involving events that occur synchronously and/or asynchronously. To effectively consume time information, recent studies have focused on designing new architectures. In this paper, we take an orthogonal but complementary approach by providing a model-agnostic vector representation for time, called \emph{\timetovec}, that can be easily imported into many existing and future architectures and improve their performances. We show on a range of models and problems that replacing the notion of time with its \timetovec\ representation improves the performance of the final model. 
\end{abstract}

\section{Introduction}
In building machine learning models, ``time'' is often an important feature.
Examples include predicting daily sales for a company based on the date (and other available features), predicting the time for a patient's next health event based on their medical history, and predicting the song a person is interested in listening to based on their listening history.
The input for problems involving time can be considered as a sequence where, rather than being identically and independently distributed (\emph{iid}), there exists a dependence across time (and/or space) among the data points.
The sequence can be either synchronous, \ie  sampled at regular intervals, or asynchronous, \ie sampled at different points in time.
In both cases, time may be an important feature.
For predicting daily sales, for instance,
it may be useful to know if it is a holiday or not.
For predicting the time for a patient's next encounter, it is important to know the (asynchronous) times of their previous visits.

Recurrent neural networks (RNNs) \cite{hochreiter1997long,cho2014learning} have achieved impressive results on a range of sequence modeling problems.
Most RNN models do not treat time itself as a feature, typically assuming that inputs are synchronous.
When time is known to be a relevant feature, it is often fed in as yet another input dimension \cite{choi2016doctor,du2016recurrent,46654}.
In practice, RNNs often fail at effectively making use of time as a feature. To help the RNN make better use of time, several researchers design hand-crafted features of time that suit their specific problem and feed those features into the RNN \cite{choi2016doctor,baytas2017patient,kwon2019retainvis}. Hand-crafting features, however, can be expensive and requires domain expertise about the problem.

Many recent studies aim at obviating the need for hand-crafting features by proposing general-purpose---as opposed to problem specific---architectures that better handle time \cite{neil2016phased,zhu2017next,mei2017neural,hu2017state,upadhyay2018deep,li2018learning}.
We follow an orthogonal but complementary approach to these recent studies by developing a general-purpose model-agnostic representation for time that can be potentially used in any architecture. 
In particular, we develop a learnable vector representation (or embedding) for time as a vector representation can be easily combined with many models or architectures. 
We call this vector representation \emph{\timetovec}. 
To validate the effectiveness of \timetovec, we conduct experiments on several (synthesized and real-world) datasets and integrate it with several architectures. 
Our main result is to show that on a range of problems and architectures that consume time, using \timetovec\ instead of the time itself offers a boost in performance.

\section{Related Work}

There is a long history of algorithms for predictive modeling in time series analysis.
They include auto-regressive techniques~\cite{akaike1969fitting} that predict future measurements in a sequence based on a window of past measurements.
Since it is not always clear how long the window of past measurements should be, hidden Markov models~\cite{rabiner1986introduction}, dynamic Bayesian networks~\cite{murphy2002dynamic}, and dynamic conditional random fields~\cite{sutton2007dynamic} use hidden states as a finite memory that can remember information arbitrarily far in the past.
These models can be seen as special cases of recurrent neural networks~\cite{hochreiter1997long}.  
They typically assume that inputs are synchronous, \ie arrive at regular time intervals, and that the underlying process is stationary with respect to time.
It is possible to aggregate asynchronous events into time-bins and to use synchronous models over the bins~\cite{lipton2016directly,anumula2018feature}.
Asynchronous events can also be directly modeled with point processes (\eg, Poisson, Cox, and Hawkes point processes)~\cite{daley2007introduction,laub2015hawkes,xiao2017wasserstein,li2018learning,xiao2018learning} and continuous time normalizing flows \cite{chen2018neural}.
Alternatively, one can also interpolate or make predictions at arbitrary time stamps with Gaussian processes~\cite{rasmussen2004gaussian} or support vector regression~\cite{drucker1997support}.  

Our goal is not to propose a new model for time series analysis, but instead to propose a representation of time in the form of a vector embedding that can be used by many models.
Vector embedding has been previously successfully used for other domains such as text \cite{mikolov2013distributed,pennington2014glove}, (knowledge) graphs \cite{grover2016node2vec,nickel2016review,kazemi2018simple}, and positions \cite{vaswani2017attention,gehring2017convolutional}. 
Our approach is related to time decomposition techniques that encode a temporal signal into a set of frequencies~\cite{cohen1995time}. However, instead of using a fixed set of frequencies as in Fourier transforms~\cite{bracewell1986fourier}, we allow the frequencies to be learned.
We take inspiration from the neural decomposition of \citet{godfrey2018neural} (and similarly \cite{gashler2016modeling}).
For time-series analysis, \citet{godfrey2018neural} decompose a 1D signal of time into several \sine\ functions and a linear function
to extrapolate (or interpolate) the given signal.
We follow a similar intuition but instead of decomposing a 1D signal of time into its components, we transform the time itself and feed its transformation into the model that is to consume the time information.
Our approach corresponds to the technique of \citet{godfrey2018neural} when applied to regression tasks in 1D signals, but it is more general since we learn a representation that can be shared across many signals and can be fed to many models for tasks beyond regression.

While there is a body of literature on designing neural networks with \sine\ activations \cite{lapedes1987nonlinear,sopena1999neural,wong2002handwritten,mingo2004fourier,liu2016multistability}, our work uses \sine\ only for transforming time; the rest of the network uses other activations. 
There is also a set of techniques that consider time as yet another feature and concatenate time (or some hand designed features of time such as log and/or inverse of delta time) with the input~\cite{choi2016doctor,li2017time,du2016recurrent,baytas2017patient,kwon2019retainvis,trivedi2017know,kumar2018learning,ma2018dynamic}. \citet{kazemi2019relational} survey several such approaches for dynamic (knowledge) graphs.
These models can directly benefit from our proposed vector embedding, Time2Vec, by concatenating Time2Vec with the input instead of their time features.
Other works~\cite{neil2016phased,zhu2017next,mei2017neural,hu2017state,upadhyay2018deep,li2018learning} propose new neural architectures that take into account time (or some features of time).
As a proof of concept, we show how \timetovec\ can be used in one of these architectures to better exploit temporal information; it can be potentially used in other architectures as well.
 
\section{Background \& Notation} \label{background-section}

We use lower-case letters to denote scalars, bold lower-case letters to denote vectors, and bold upper-case letters to denote matrices. We represent the $i^{th}$ element of the vector $\vctr{r}$ as $\vctr{r}[i]$. For two vectors $\vctr{r}$ and $\vctr{s}$, we use $[\vctr{r};\vctr{s}]$ to represent their concatenation and $\vctr{r}\odot\vctr{s}$ to represent element-wise (Hadamard) multiplication of the two vectors. Throughout the paper, we use $\tau$ to represent a scalar notion of time (\eg, absolute time, time from start, time from the last event, etc.) and $\vctr{\tau}$ to represent a vector of time features.

Long Short Term Memory (LSTM) \cite{hochreiter1997long} is considered one of the most successful RNN architectures for sequence modeling.
A formulation of the original LSTM model and a variant of it based on peepholes \cite{gers2000recurrent} is presented in Appendix C.
When time is a relevant feature, the easiest way to handle time is to consider it as just another feature (or extract some engineered features from it), concatenate the time features with the input, and use the standard LSTM model (or some other sequence model) \cite{choi2016doctor,du2016recurrent,46654}. In this paper, we call this model \emph{LSTM+T}. Another way of handling time is by changing the formulation of the standard LSTM.
\citet{zhu2017next} developed one such formulation, named \emph{TimeLSTM}, by adding time gates to the architecture of the LSTM with peepholes. They proposed three architectures namely \emph{TLSTM1}, \emph{TLSTM2}, \emph{TLSTM3}. A description of \emph{TLSTM1} and \emph{TLSTM3} can be found in Appendix C (we skipped TLSTM2 as it is quite similar to TLSTM3).
 
\section{Time2Vec} \label{sec:time2vec}
In designing a representation for time, we identify three important properties: 1- capturing both periodic and non-periodic patterns, 2- being invariant to time rescaling, and 3- being simple enough so it can be combined with many models. In what follows, we provide more detail on these properties.

\textbf{Periodicity:} In many scenarios, some events occur periodically. The amount of sales of a store, for instance, may be higher on weekends or holidays.
Weather condition usually follows a periodic pattern over different seasons \cite{gashler2016modeling}. Some notes in a piano piece usually repeat periodically \cite{hu2017state}. Some other events may be non-periodic but only happen after a point in time and/or become more probable as time proceeds. For instance, some diseases are more likely for older ages. 
Such periodic and non-periodic patterns distinguish time from other features calling for better treatment and a better representation of time. In particular, it is important to use a representation that enables capturing periodic and non-periodic patterns.

\textbf{Invariance to Time Rescaling:} Since time can be measured in different scales (\eg, days, hours, seconds, etc.), another important property of a representation for time is invariance to time rescaling (see, \eg, \cite{tallec2018iclr}). A class $\mathcal{C}$ of models is invariant to time rescaling if for any model $\mathcal{M}_1\in\mathcal{C}$ and any scalar $\alpha > 0$, there exists a model $\mathcal{M}_2\in\mathcal{C}$ that behaves on $\alpha\tau$ ($\tau$ scaled by $\alpha$) in the same way $\mathcal{M}_1$ behaves on original $\tau$s.

\textbf{Simplicity:} A representation for time should be easily consumable by different models and architectures. A matrix representation, for instance, may be difficult to consume as it cannot be easily appended with the other inputs. 

\textbf{\timetovec:} We propose \emph{\timetovec}, a representation for time which has the three identified properties.  For a given scalar notion of time $\tau$, \timetovec\ of $\tau$, denoted as $\ttov(\tau)$, is a vector of size $k+1$ defined as follows:  
\begin{equation}
    \label{eq:t2vec_def}
  \ttov(\tau)[i]=\begin{cases}
    \omega_i \tau + \varphi_i, & \text{if~~$i=0$}. \\
    \periodicF{(\omega_i \tau + \varphi_i)}, & \text{if~~$1\leq i \leq k$}.
  \end{cases}
\end{equation} 
where $\ttov(\tau)[i]$ is the $i^{th}$ element of $\ttov(\tau)$, \periodicF\ is a periodic activation function, and $\omega_i$s and $\varphi_i$s are learnable parameters. Given the prevalence of vector representations for different tasks, a vector representation for time makes it easily consumable by different architectures.
We chose \periodicF\ to be the \sine\ function in our experiments\footnote{Using cosine (and some other similar activations) results in an equivalent representation.} but we do experiments with other periodic activations as well.
When $\mathcal{F}=\sin$, for $1 \leq i \leq k$, $\omega_i$ and $\varphi_i$ are the frequency and the phase-shift of the \sine\ function.  

The period of $\sin{(\omega_i \tau + \varphi_i)}$ is $\frac{2\pi}{\omega_i}$, \ie it has the same value for $\tau$ and $\tau+\frac{2\pi}{\omega_i}$. Therefore, a \sine\ function helps  capture periodic behaviors without the need for feature engineering. For instance, a \sine\ function $\sin{(\omega \tau+\varphi)}$ with $\omega=\frac{2\pi}{7}$ repeats every $7$ days (assuming $\tau$ indicates days) and can be potentially used to model weekly patterns. Furthermore, unlike other basis functions which may show strange behaviors for extrapolation (see, \eg, \cite{poole2014population}), \sine\ functions are expected to work well for extrapolating to future and out of sample data \cite{vaswani2017attention}. The linear term represents the progression of time and can be used for capturing non-periodic patterns in the input that depend on time. 
Proposition~\ref{invariance-prop} establishes the invariance of \timetovec\ to time rescaling. The proof is in Appendix D.

\begin{proposition} \label{invariance-prop}
\timetovec\ is invariant to time rescaling.
\end{proposition}

The use of \sine\ functions is inspired in part by \citet{vaswani2017attention}'s positional encoding. Consider a sequence of items (\eg, a sequence of words) $\{I_1, I_2, \dots, I_N\}$ and a vector representation $\vctr{v}_{I_j}\in\mathbb{R}^d$ for the $j^{th}$ item $I_j$ in the sequence. \citet{vaswani2017attention} added $\sin{(j/10000^{k/d})}$ to $\vctr{v}_{I_j}[k]$ if $k$ is even and $\sin{(j/10000^{k/d} + \pi/2)}$ if $k$ is odd so that the resulting vector includes information about the position of the item in the sequence. These \sine\ functions are called the positional encoding. Intuitively, positions can be considered as the times and the items can be considered as the events happening at that time. Thus, \timetovec\ can be considered as representing continuous time, instead of discrete positions, using \sine\ functions. The \sine\ functions in \timetovec\ also enable capturing periodic behaviors which is not a goal in positional encoding. We feed \timetovec\ as an input to the model (or to some gate in the model) instead of adding it to other vector representations. Unlike positional encoding, we show in our experiments that learning the frequencies and phase-shifts of \sine\ functions in \timetovec\ result in better performance compared to fixing them.

\section{Experiments \& Results}
We design experiments to answer the following questions: \textbf{Q1:} is \timetovec\ a good representation for time?, \textbf{Q2:} can \timetovec\ be used in other architectures and improve their performance?, \textbf{Q3:} what do the \sine\ functions learn?, \textbf{Q4:} can we obtain similar results using non-periodic activation functions for Eq.~\eqref{eq:t2vec_def} instead of periodic ones?, and \textbf{Q5:} is there value in learning the \sine\ frequencies or can they be fixed (\eg, to equally-spaced values as in Fourier sine series or  exponentially-decaying values as in \citet{vaswani2017attention}'s positional encoding)?
We use the following datasets:

\textbf{1) Synthesized data:} We create a toy dataset to use for explanatory experiments. The inputs in this dataset are the integers between $1$ and $365$. Input integers that are multiples of $7$ belong to class one and the other integers belong to class two. 
The first 75\% is used for training and the last 25\%  for testing. This dataset is inspired by the periodic patterns (\eg, weekly or monthly) that often exist in daily-collected data; the input integers can be considered as the days.

\textbf{2) \emnist:} Sequential (event-based) MNIST is a common benchmark in sequence modeling literature (see, \eg, \cite{bellec2018long,campos2017skip,fatahi2016evt_mnist}).
We create a sequential event-based version of MNIST by flattening the images and recording the position of the pixels whose intensities are larger than a threshold ($0.9$ in our experiment).
Following this transformation, each image will be represented as an array of increasing numbers such as $[t_1, t_2, t_3, \dots, t_m]$. We consider these values as the event times and use them to classify the images. As in other sequence modeling works, our aim in building this dataset is not to beat the state-of-the-art on the MNIST dataset; our aim is to provide a dataset where the only input is time and different representations for time can be compared when extraneous variables (confounders) are eliminated as much as possible.

\textbf{3) \ntdigits}~\cite{anumula2018feature}: The dataset includes audio spikes of the TIDIGITS spoken digit dataset~\cite{leonard1993tidigits} recorded by the binaural 64-channel silicon cochlea sensor. Each sample is a sequence of $(t,c)$ tuples where $t$ represents time and $c$ denotes the index of active frequency channel at time $t$. The labels are sequences of 1 to 7 connected digits with a vocabulary consisting of 11 digits (i.e. ``zero'' to ``nine'' plus ``oh'') and the goal is to classify the spoken digit based on the given sequence of active channels. We use the reduced version of the dataset where only the single digit samples are used for training and testing. The reduced dataset has a total of 2,464 training and 2,486 test samples. 

\textbf{4) Stack Overflow (SOF):} This dataset contains sequences of badges obtained by stack overflow users and the timestamps at which the badges were obtained\footnote{\url{https://archive.org/details/stackexchange}}. We used the subset released by \citet{du2016recurrent} containing $\sim 6K$ users, $22$ event types (badges), and $\sim 480K$ events. Given a sequence $[(b_1^u,t_1^u),(b_2^u,t_2^u),...,(b_n^u,t_n^u)]$ for each user $u$ where $b_i^u$ is the badge id and $t_i^u$ is the timestamp when $u$ received this badge id, the task is to predict the badge the user will obtain at time $t_{k+1}^u$.

\textbf{5) \lastfm:} This dataset contains a history of listening habits for \lastfm\ users \cite{Celma:Springer2010}.
We used the code released by \citet{zhu2017next} to pre-process the data.
The dataset contains $\sim 1K$ users, 5000 event types (songs), and $\sim 819K$ events.
The prediction problem is similar to the SOF dataset but with dynamic updating (see, \cite{zhu2017next} for details).

\textbf{6) \culike:} This dataset contains data about what and when a user posted on citeulike website\footnote{\url{http://www.citeulike.org/}}. 
The original dataset has about 8000 samples. Similar to \lastfm, we used the pre-processing used by \citet{zhu2017next} to select $\sim1.6K$ sequences with 5000 event types (papers) and $\sim36K$ events. 
The task for this dataset is similar to that for \lastfm. 

\begin{figure}
   \centering
   \subfloat[\emnist]{%
   \includegraphics[width=0.31\textwidth]{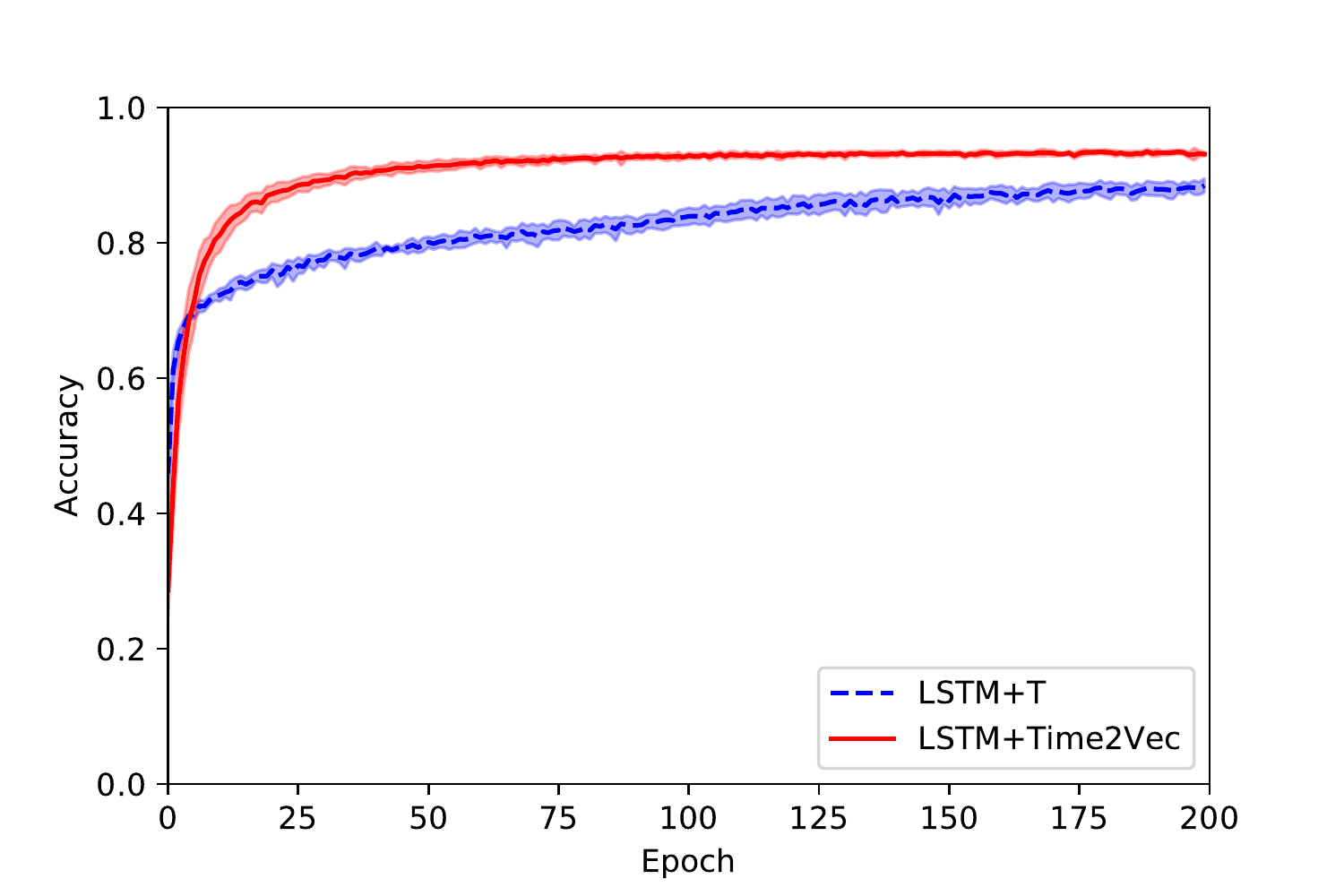}}
~~~~\hspace*{0cm}
   \subfloat[Raw \ntdigits]{%
   \includegraphics[width=0.31\textwidth]{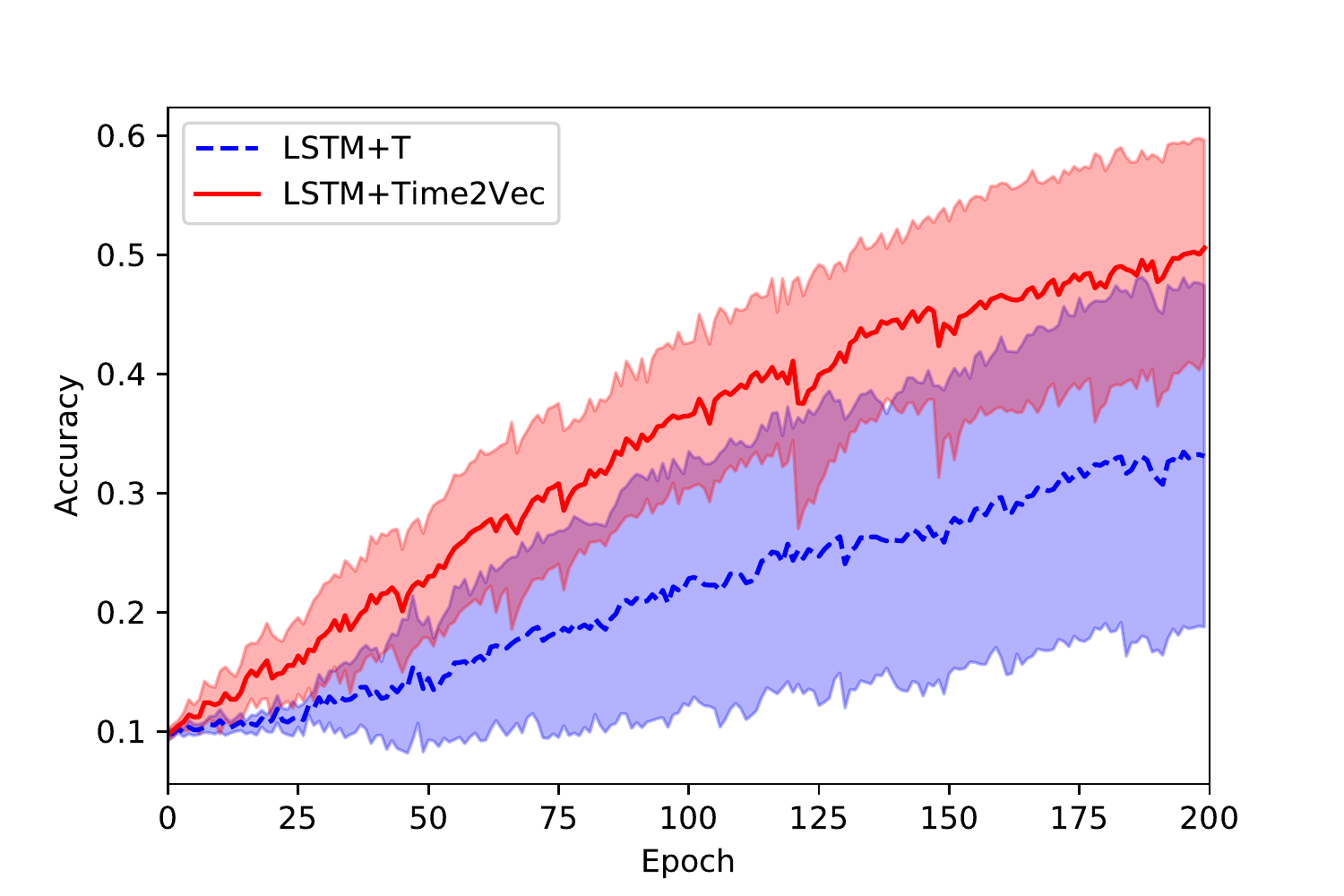}} 
   ~~~~\hspace*{0cm}
   \subfloat[Stack Overflow]{%
   \includegraphics[width=0.31\textwidth]{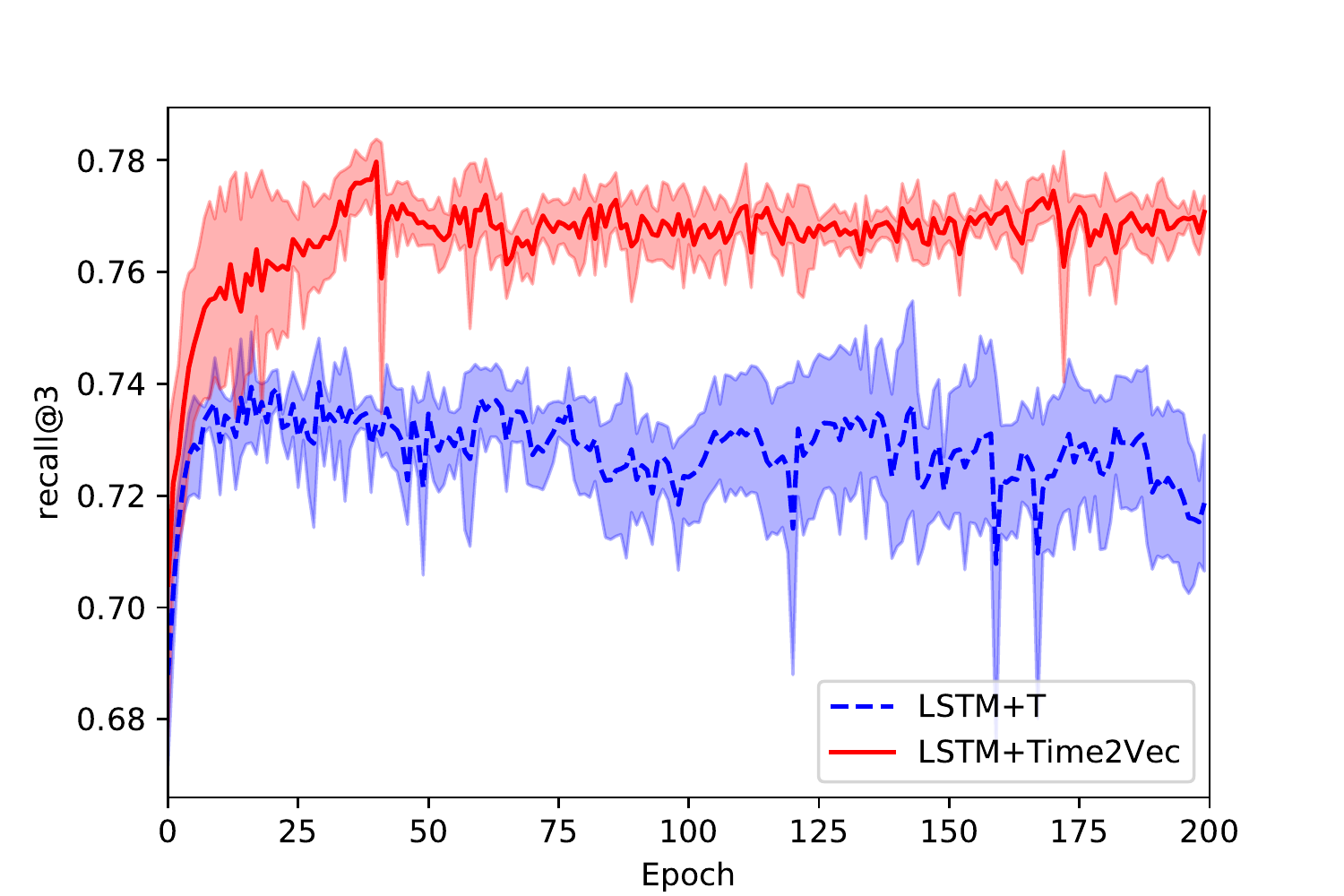}} 
   \\ \vspace*{-0.3cm}
   \subfloat[\lastfm]{%
   \includegraphics[width=0.35\textwidth]{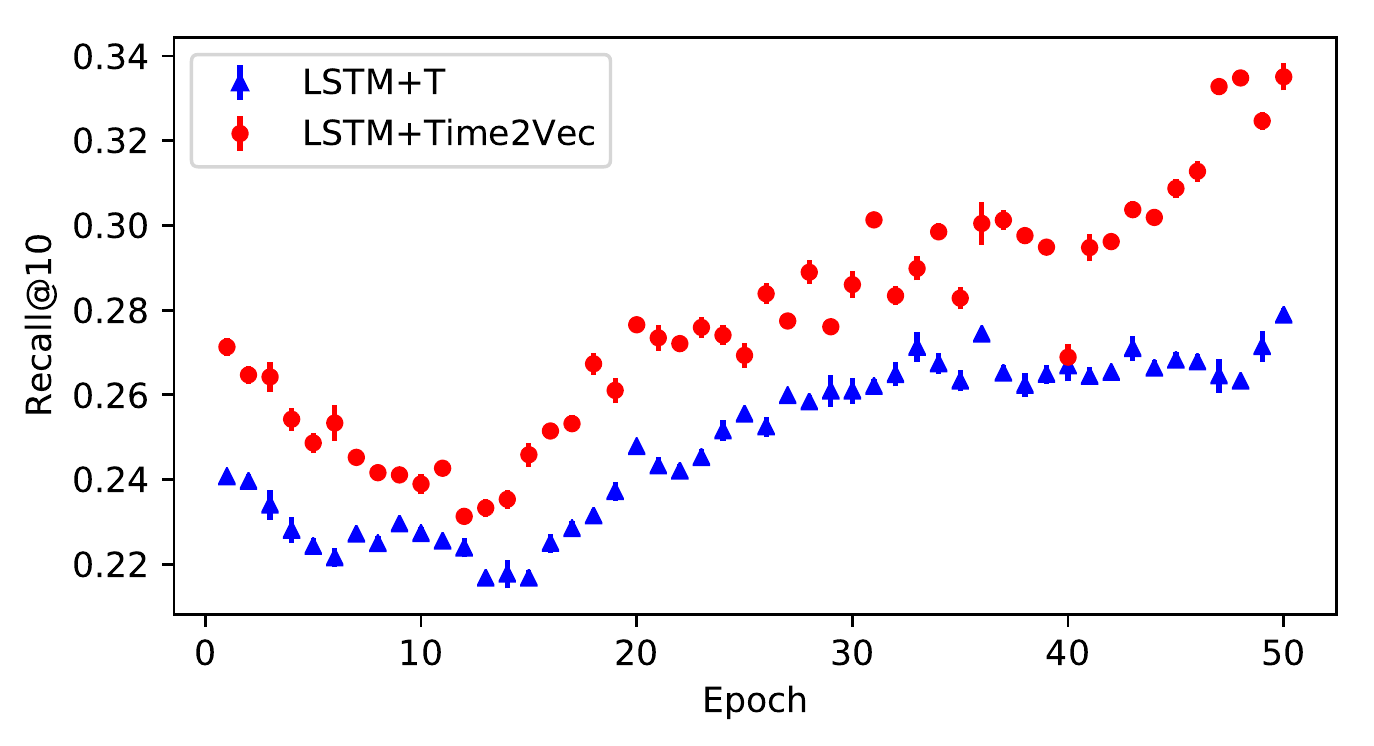}}
~~~~\hspace*{0cm}
   \subfloat[\culike]{%
   \includegraphics[width=0.35\textwidth]{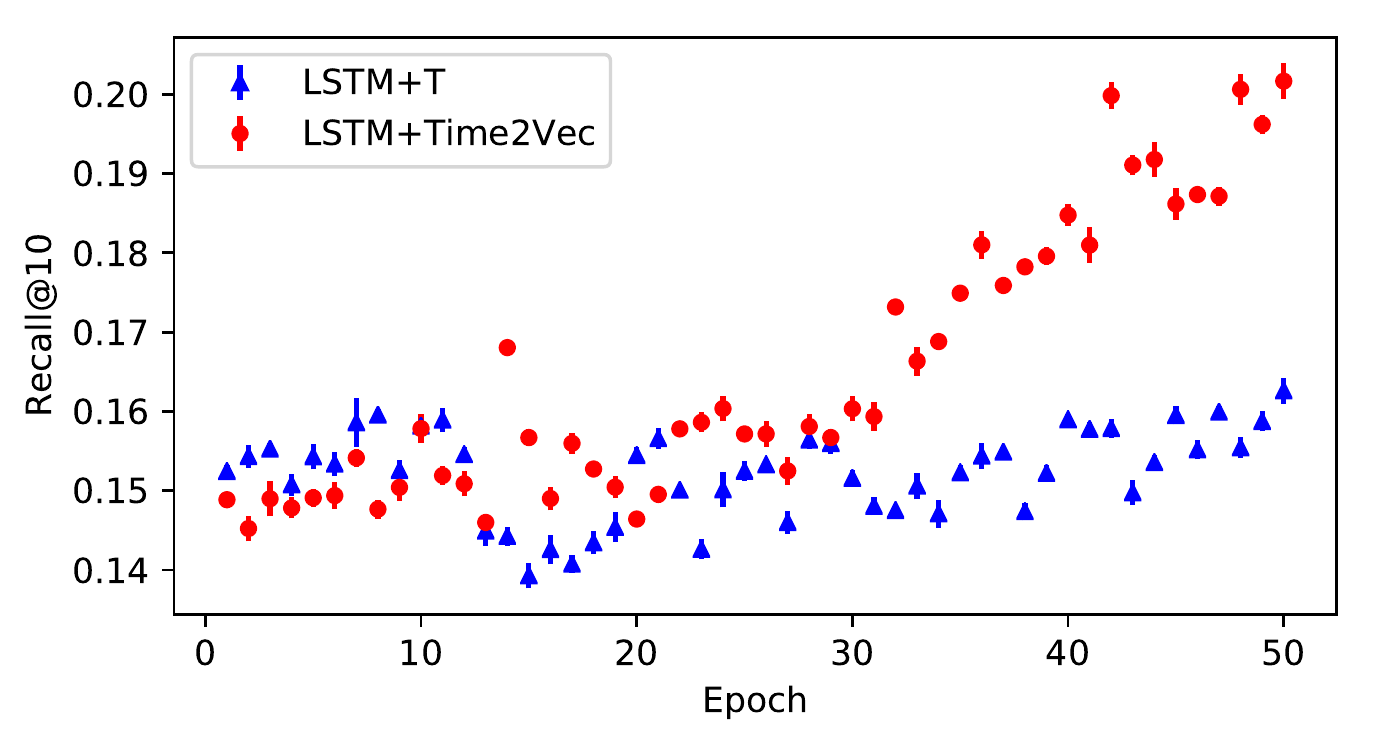}}

   \caption{%
   \label{fig:comparisons} %
   Comparing LSTM+T and LSTM+\timetovec\ on several datasets.}
\end{figure}

\textbf{Measures:} 
For classification tasks, we report \emph{accuracy} corresponding to the percentage of correctly classified examples.
For recommendation tasks, we report \emph{Recall@q}\ and \emph{MRR@q}. Following \citet{zhu2017next}, to generate a recommendation list, we sample $k-1$ random items and add the correct item to the sampled list resulting in a list of $k$ items. Then our model ranks these $k$ items. Looking only at the top ten recommendations, Recall@q corresponds to the percentage of recommendation lists where the correct item is in the top $q$; MRR@q (reported in Appendix B) corresponds to the mean of the inverses of the rankings of the correct items where the inverse rank is considered $0$ if the item does not appear in top $q$ recommendations. For \lastfm\ and \culike, following \citet{zhu2017next} we report Recall@10 and MRR@10. For SOF, we report Recall@3 and MRR as there are only $22$ event types and Recall@10 and MRR@10 are not informative enough. The detail of the implementations is presented in Appendix A.

\subsection{On the effectiveness of \timetovec} \label{comparison-results}
Fig.~\ref{fig:comparisons} represents the obtained results of comparing \emph{LSTM+\timetovec} with \emph{LSTM+T} on several datasets with different properties and statistics.
It can be observed that on all datasets, replacing time with \timetovec\ improves the performance in most cases and never deteriorates it; in many cases, LSTM+\timetovec\ performs consistently better than LSTM+T. 
\citet{anumula2018feature} mention that LSTM+T fails on \ntdigits\ as the dataset contains very long sequences. By feeding better features to the LSTM rather than relying on the LSTM to extract them, \timetovec\ helps better optimize the LSTM and offers higher accuracy (and lower variance) compared to LSTM+T.
Besides \ntdigits, SOF also contains somewhat long sequences and long time horizons.
The results on these two datasets indicate that \timetovec\ can be effective for datasets with long sequences and time horizons. 
From the results obtained in this subsection, it can be observed that \timetovec\ is indeed an effective representation of time thus answering \textbf{Q1} positively.

To verify if \timetovec\ can be integrated with other architectures and improve their performance, we integrate it with TLSTM1 and TLSTM3, two recent and powerful models for handling asynchronous events. To use \timetovec\ in these two architectures, we replaced their notion $\tau$ of time with $\ttov(\tau)$ and replaced the vectors getting multiplied to $\tau$ with matrices accordingly. The updated formulations are presented in Appendix C.
The obtained results in Fig.~\ref{fig:TLSTM-results} for TLSTM1 and TLSTM3 on \lastfm\ and \culike\ demonstrates that replacing time with \timetovec\ for both TLSTM1 and TLSTM3 improves the performance thus answering \textbf{Q2} affirmatively positively.

\subsection{What does \timetovec\ learn?} \label{syn-exp-section}
\begin{figure}[t]
   \centering
   \subfloat[A weighted sum of the sinusoids in \timetovec\ oscillating every $7$ days.]{%
   \includegraphics[width=0.40\textwidth]{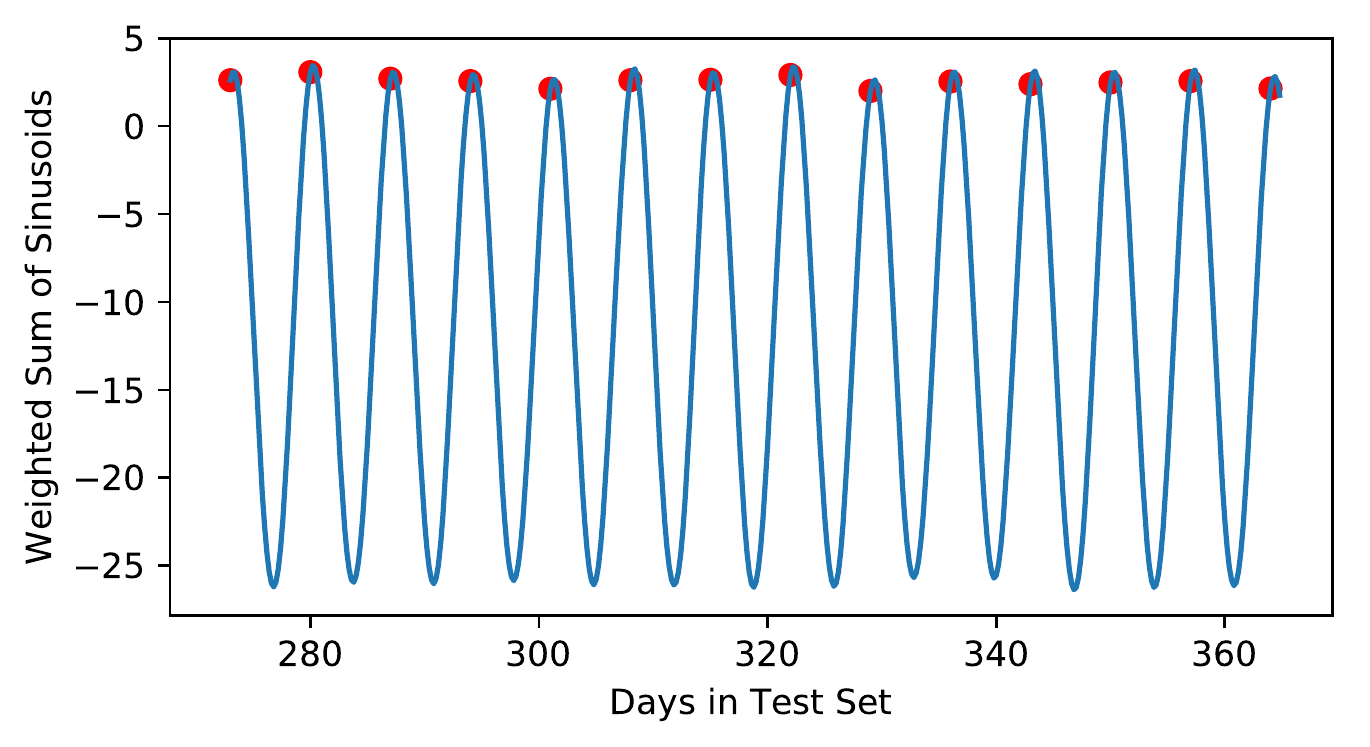}}
~~~~\hspace*{1cm}
   \subfloat[A weighted sum of the sinusoids in \timetovec\ oscillating every $14$ days.]{%
   \includegraphics[width=0.40\textwidth]{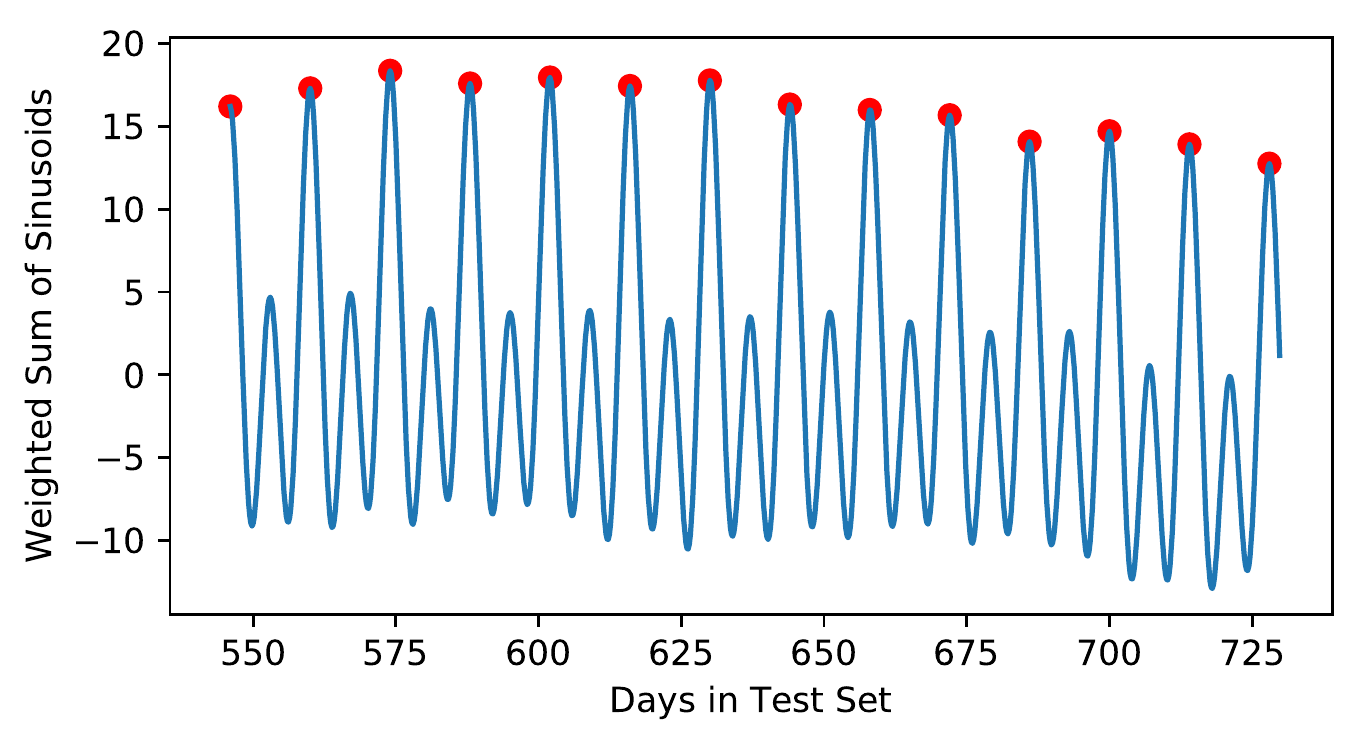}} %

   \caption{%
   \label{synthesized-fig} %
   The models learned for our synthesized dataset explained in Subsection~\ref{syn-exp-section} before the final activation. The red dots represent the points to be classified as $1$.}
\end{figure}

\begin{figure*}[b!]
  \centering
\includegraphics[width=0.59\textwidth]{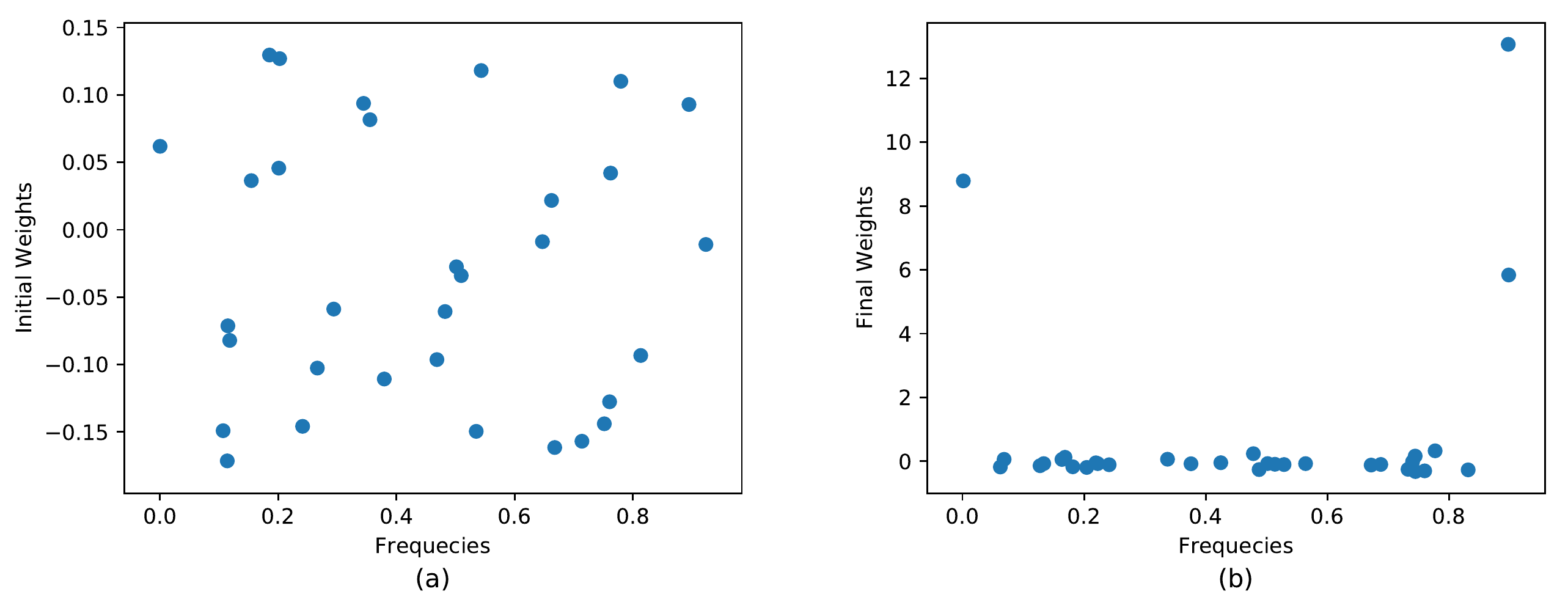}
  \caption{(a) Initial vs. (b) learned weights and frequencies for our synthesized dataset.}
  \label{init_vs_learned_weights-fig}
\end{figure*}

To answer \textbf{Q3}, we trained a model on our synthesized dataset where the input integer (day) is used as the time for \timetovec\ and a fully connected layer is used on top of the \timetovec\ to predict the class. That is, the probability of one of the classes is a sigmoid of a weighted sum of the \timetovec\ elements.
Fig.~\ref{synthesized-fig}~(a) shows a the learned function for the days in the test set where the weights, frequencies and phase-shifts are learned from the data. The red dots on the figure represent multiples of $7$. It can be observed that \timetovec\ successfully learns the correct period and oscillates every $7$ days. The phase-shifts have been learned in a way that all multiples of $7$ are placed on the positive
peaks of the signal to facilitate separating them from the other days.
Looking at the learned frequency and phase-shift for the \sine\ functions across several runs, we observed that in many runs one of the main \sine\ functions has a frequency around $0.898\approx\frac{2\pi}{7}$ and a phase-shift around $1.56\approx\frac{\pi}{2}$, thus learning to oscillate every $7$ days and shifting by $\frac{\pi}{2}$ to make sure multiples of $7$ end up at the peaks of the signal. Fig.~\ref{init_vs_learned_weights-fig} shows the initial and learned sine frequencies for one run. It can be viewed that at the beginning, the weights and frequencies are random numbers. But after training, only the desired frequency ($\frac{2\pi}{7}$) has a high weight (and the $0$ frequency which gets subsumed into the bias).
The model perfectly classifies the examples in the test set which represents the \sine\ functions in \timetovec\ can be used effectively for extrapolation and out of sample times assuming that the test set follows similar periodic patterns as the train set\footnote{Replacing \sine\ with a non-periodic activation function resulted in always predicting the majority class.}. We added some noise to our labels by flipping $5\%$ of the labels selected at random and observed a similar performance in most runs.

To test invariance to time rescaling, we multiplied the inputs by $2$ and observed that in many runs, the frequency of one of the main \sine\ functions was around $0.448\approx\frac{2\pi}{2*7}$ thus oscillating every $14$ days. 
An example of a combination of signals learned to oscillate every $14$ days is in Fig.~\ref{synthesized-fig}~(b). 

\begin{figure}
   \centering
   \subfloat[TLSTM1, \lastfm]{%
   \includegraphics[width=0.35\textwidth]{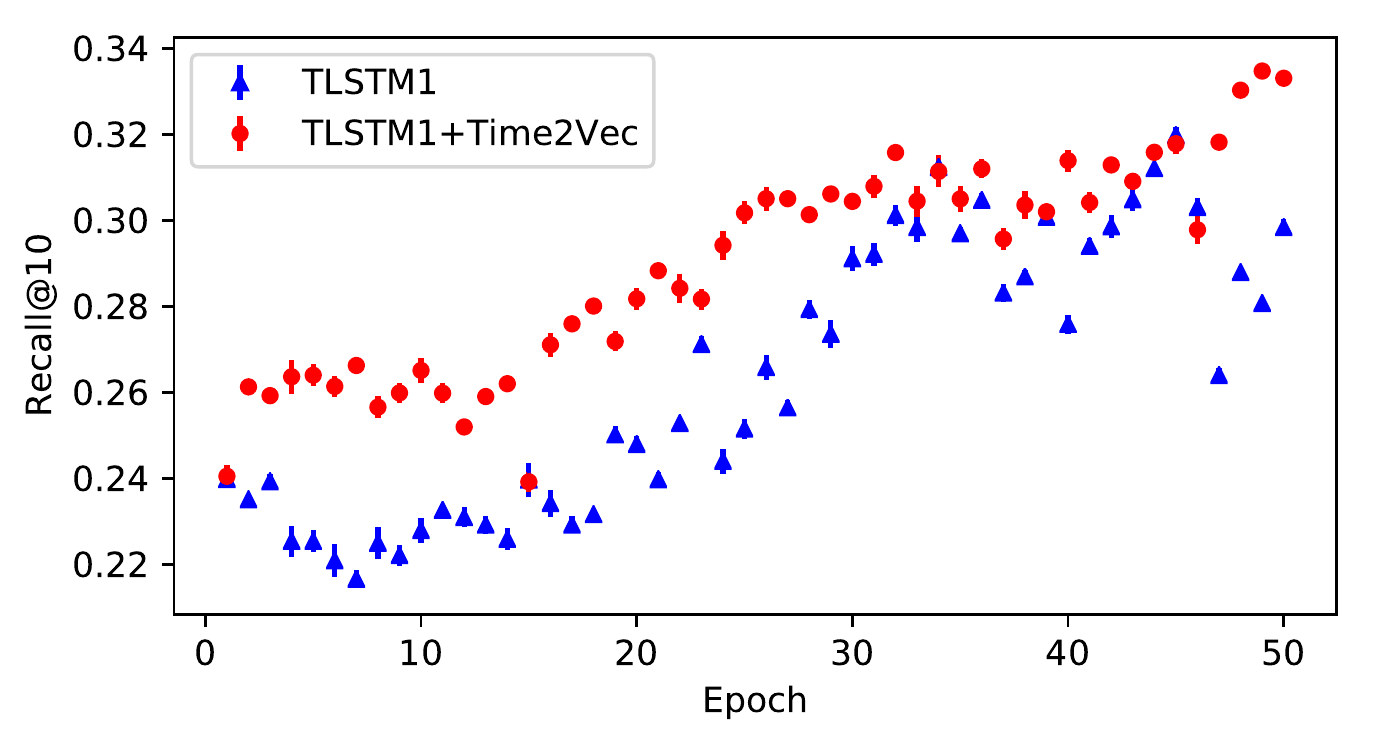}}
~~~~\hspace*{1cm}
   \subfloat[TLSTM1, \culike]{%
   \includegraphics[width=0.35\textwidth]{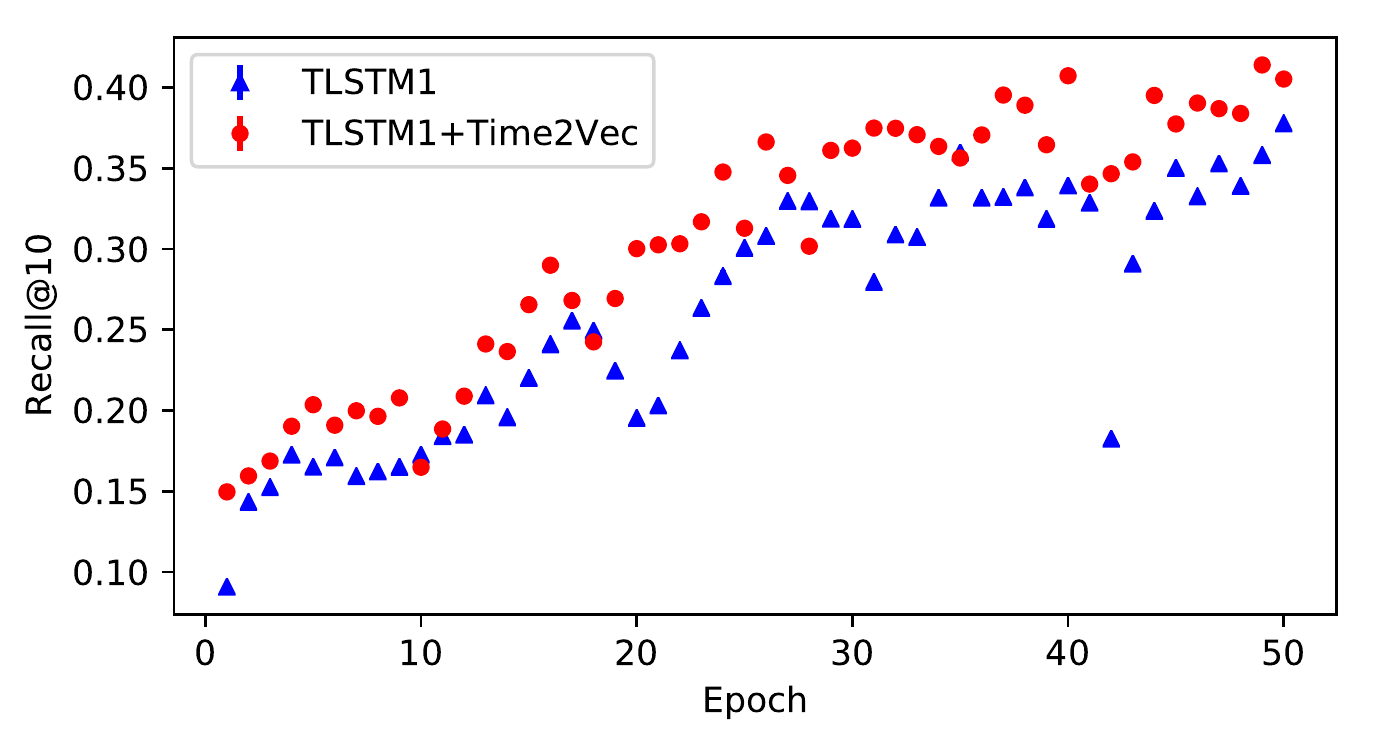}} 
   \\ \vspace*{-0.3cm}
   \subfloat[TLSTM3, \lastfm]{%
   \includegraphics[width=0.35\textwidth]{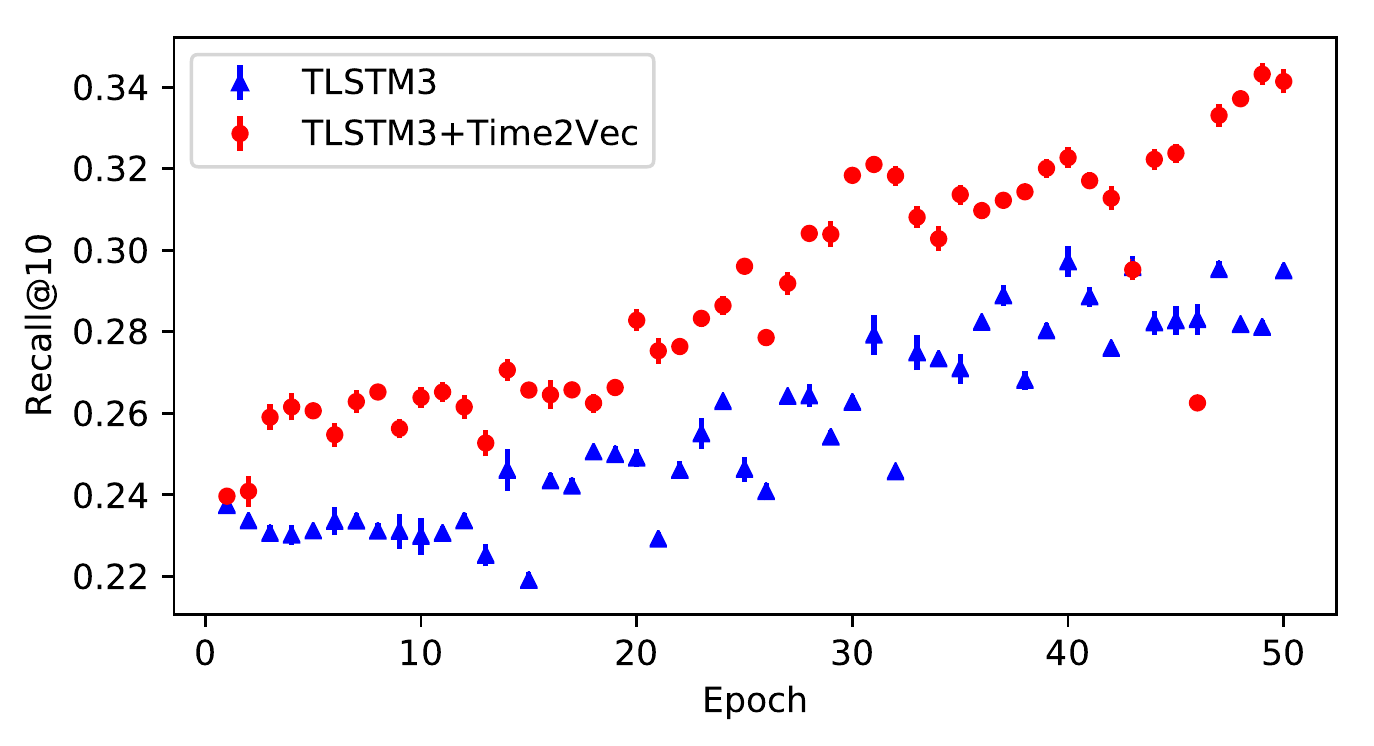}}
~~~~\hspace*{1cm}
   \subfloat[TLSTM3, \culike]{%
   \includegraphics[width=0.35\textwidth]{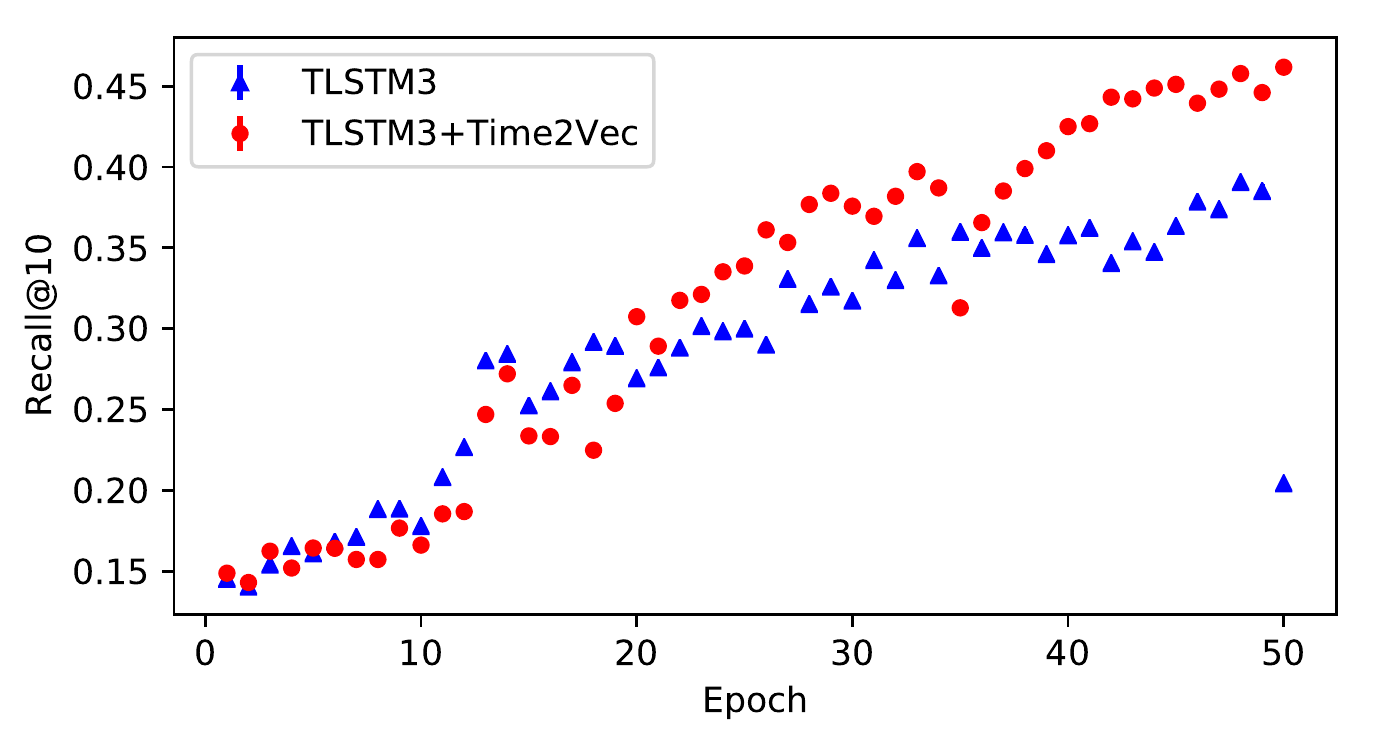}}

   \caption{%
   \label{fig:TLSTM-results} %
   Comparing TLSTM1 and TLSTM3 on \lastfm\ and \culike\ in terms of \recallten\ with and without \timetovec.
   }
\end{figure}

\subsection{Other activation functions}
\label{other-activations-subsection}
To answer \textbf{Q4}, we repeated the experiment on \emnist\ in Section~\ref{comparison-results} when using activation functions other than \sine\ including non-periodic functions such as \sigmoid, \tanhyp, and rectified linear units (\relu) \cite{nair2010rectified}, and periodic activation functions such as \emph{mod} and \emph{triangle}. We fixed the length of the \timetovec\ to $64+1$, \ie $64$ units with a non-linear transformation and $1$ unit with a linear transformation. From the results shown in Fig.~\ref{fig:activation-fix-freq}(a), it can be observed that the periodic activation functions (\sine, mod, and triangle) outperform the non-periodic ones. 
Other than not being able to capture periodic behaviors, we believe one of the main reasons why these non-periodic activation functions do not perform well is because as time goes forward and becomes larger, \sigmoid\ and \tanhyp\ saturate and \relu\ either goes to zero or explodes. 

Among periodic activation functions, \sine\ outperforms the other two.
The comparative effectiveness of the \sine\ functions in processing time-series data with periodic behavior is unsurprising; An analogy can be drawn to the concept of Fourier series, which is a well established method of representing periodic signals.
From this perspective, we can interpret the output of \timetovec\ as representing periodic features which force the  model to learn periodic attributes that are relevant to the task under consideration.
This can be shown mathematically by expanding out the output of the first layer of any sequence model under consideration. Generally, the output of the first layer, after the \timetovec\ transformation and before applying a non-linear activation, has the following form $\vctr{a} = \mtrx{\Theta}\ttov{\left(\tau\right)}$,
where $\mtrx{\Theta}$ is the first layer weights matrix having components  $\left[\mtrx{\Theta}\right]_{i,j} = \theta_{i,j}$. It follows directly from Eq.~(\ref{eq:t2vec_def}) that $\vctr{a}(\tau,k)[i] = \theta_{i,0}(\omega_{0}\tau+\varphi_{0}) + \sum_{j=1}^{k} \theta_{i,j} \sin\left( \omega_i \tau + \varphi_i\right)$, where $k-1$ is the number of \sine\ functions used in the \timetovec\ transformation. Hence, the first layer has transformed time $\tau$ into $n$ (output dimension of first layer) distinct features. 
The linear term $\theta_{i,0}(\omega_{0}\tau+\varphi_{0})$ can model non-periodic components and helps with extrapolation \cite{godfrey2018neural}.
The second term (the sum of the weighted \sine\ functions) can be used to model the periodic behavior of the $\vctr{a}[i]$ features. Each component $\vctr{a}[i]$ of the attribute vector will latch on to different relevant signals with different underlying frequencies such that each component $\vctr{a}[i]$ can be viewed as representing distinct temporal attributes (having both a periodic and non-periodic part). The function can model a real Fourier signal when the frequencies $\omega_i$ of the \sine\ functions are integer multiples of a base (first harmonic) frequency (and $\omega_{k+1}=0$). However, we show in Section~\ref{fixed-vs-learned-subsection} that learning the frequencies results in better generalization.

\subsection{Fixed frequencies and phase-shifts} \label{fixed-vs-learned-subsection}
According to Fourier \sine\ series, any function can be closely approximated using \sine\ functions with equally-spaced frequencies. \citet{vaswani2017attention} mention that learning sine frequencies and phase-shifts for their positional encoding gives the same performance as fixing frequencies to exponentially-decaying values and phase-shifts to $0$ and $\frac{\pi}{2}$.
This raises the question of whether learning the \sine\ frequencies and phase-shifts of \timetovec\ from data offer any advantage compared to fixing them. To answer this question, we compare three models on \emnist\ when using \timetovec\ of length $16+1$: 1- fixing $\ttov(\tau)[n]$ to $\sin{(\frac{2\pi n}{16})}$ for $n \leq 16$, 2- fixing the frequencies and phase shifts according to \citet{vaswani2017attention}'s positional encoding, and 3- learning the frequencies and phase-shifts from the data. Fig.~\ref{fig:activation-fix-freq}(b) represents our obtained results. The obtained results in Fig.~\ref{fig:activation-fix-freq}(b) show that learning the frequencies and phase-shifts rather than fixing them helps improve the performance of the model and provides the answer to \textbf{Q5}.

\subsection{The use of periodicity in \sine\ functions}
It has been argued that when \sine\ is used as the activation function, only a monotonically increasing (or decreasing) part of it is used and the periodic part is ignored \cite{taming2017}. When we use \timetovec, however, the periodicity of the \sine\ functions are also being used and seem to be key to the effectiveness of the \timetovec\ representation. Fig.~\ref{fig:activation-fix-freq}(c) shows some statistics on the frequencies learned for \emnist\ where we count the number of learned frequencies that fall within intervals of lengths $0.1$ centered at $[0.05, 0.15, \dots, 0.95]$ (all learned frequencies are between $0$ and $1$). The figure contains two peaks at $0.35$ and $0.85$. Since the input to the \sine\ functions for this problem can have a maximum value of $784$ (number of pixels in an image), \sine\ functions with frequencies around $0.35$ and $0.85$ finish (almost) $44$ and $106$ full periods. The smallest learned frequency is $0.029$ which finishes (almost) $3.6$ full periods. These values indicate that the model is indeed using the periodicity of the \sine\ functions, not just a monotonically increasing (or decreasing) part of them. 

\subsection{Why capture non-periodic patterns?}
\begin{figure}
   \centering
   \subfloat[]{%
   \includegraphics[width=0.31\textwidth]{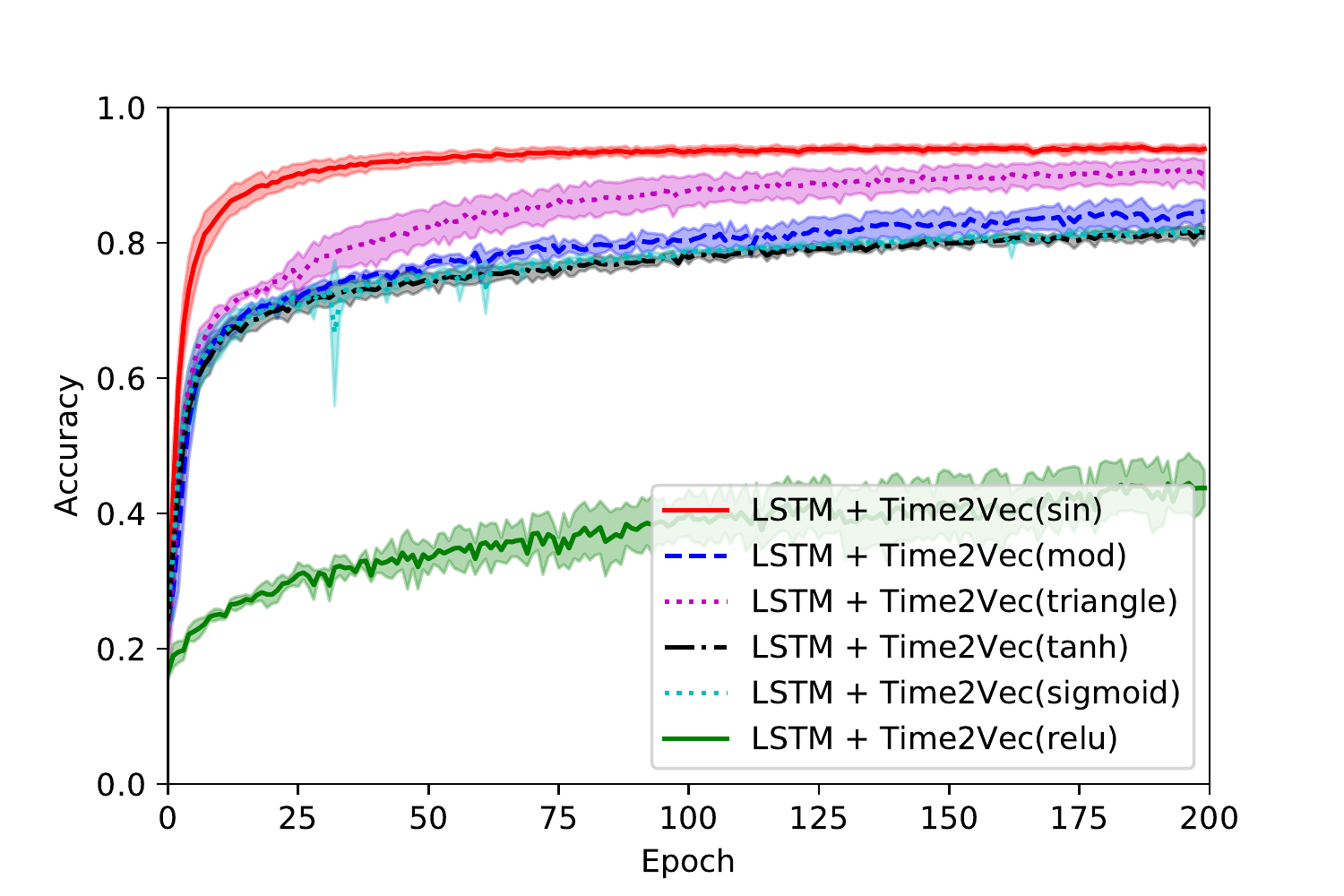}}
   ~~~~\hspace*{0.0cm}
   \subfloat[]{%
   \includegraphics[width=0.31\textwidth]{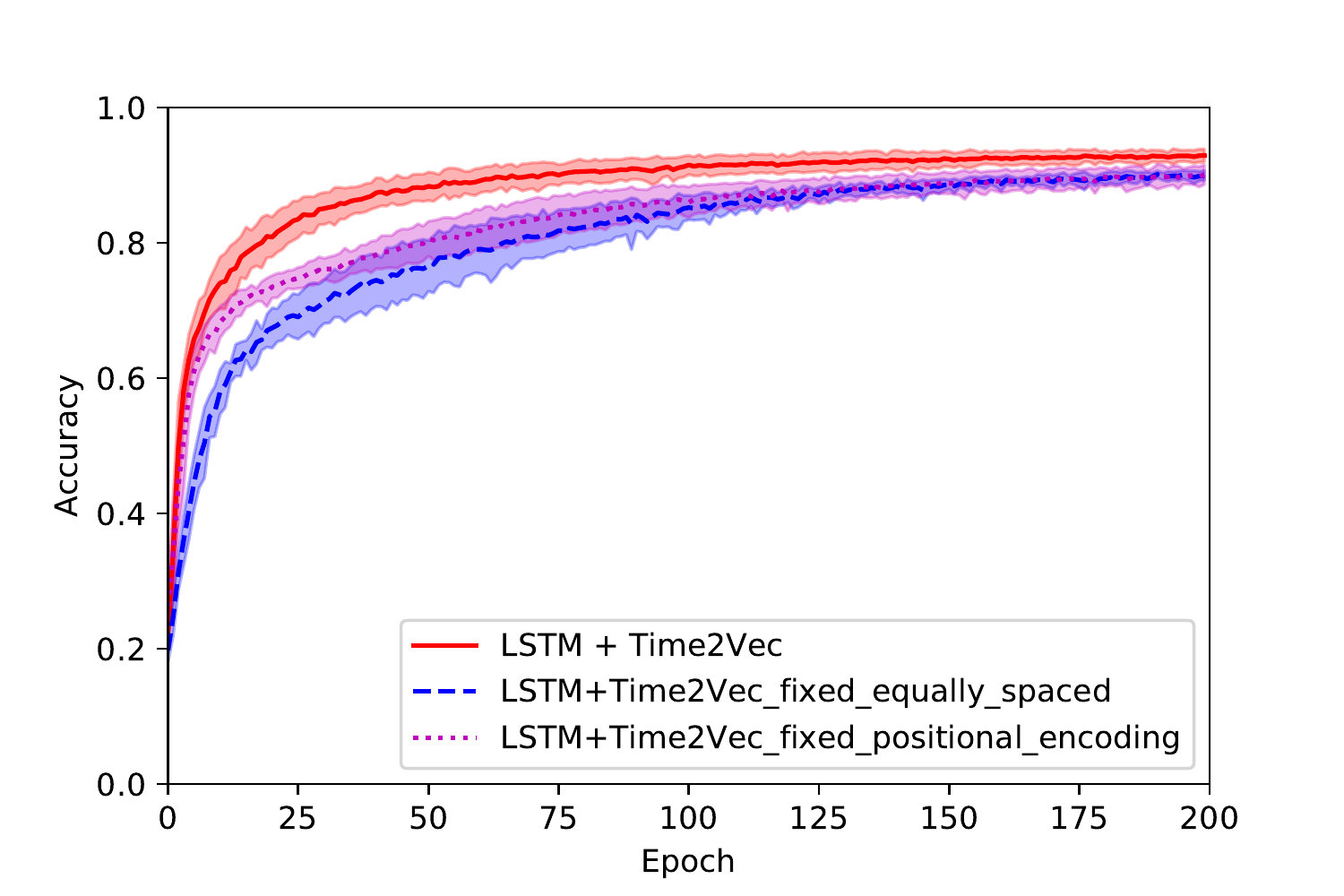}} %
   ~~~~\\ \vspace*{-0.1cm}
   \subfloat[]{%
   \includegraphics[width=0.35\textwidth]{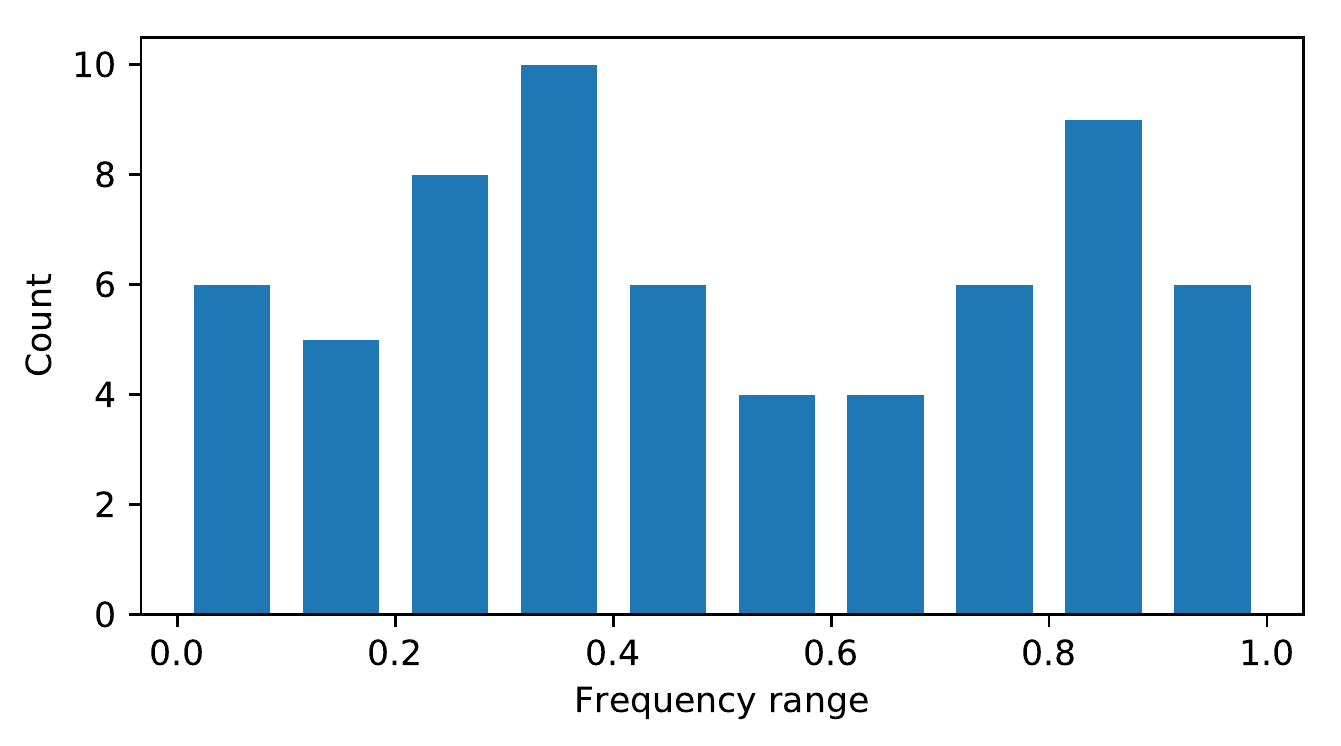}} %
   ~~~~\hspace*{0.0cm}
   \subfloat[]{%
   \includegraphics[width=0.35\textwidth]{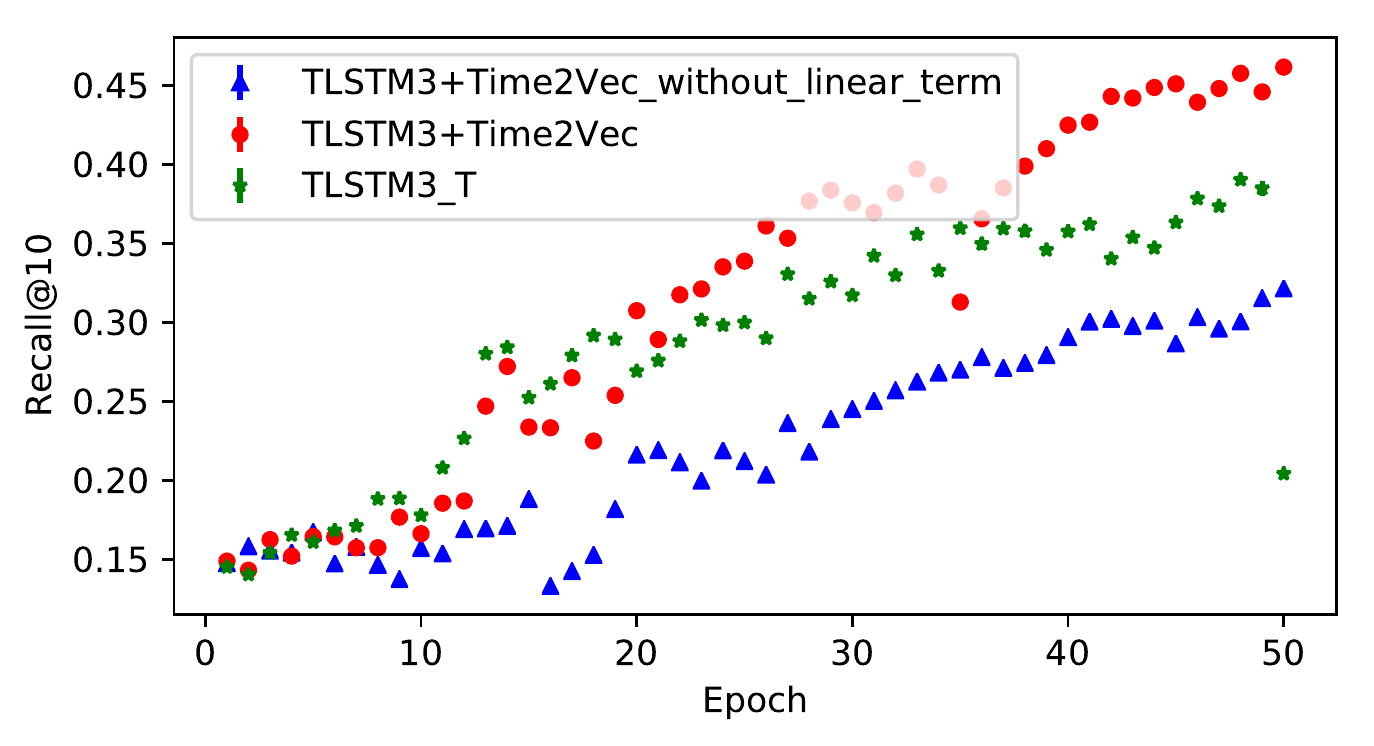}}
   \caption{%
   \label{fig:activation-fix-freq} %
  An ablation study of several components in \timetovec. (a) Comparing different activation functions for \timetovec\ on \emnist. \sigmoid\ and \tanhyp\ almost overlap. (b) Comparing frequencies fixed to equally-spaced values, frequencies fixed according to positional encoding \cite{vaswani2017attention}, and learned frequencies on \emnist. (c) A histogram of the frequencies learned in \timetovec\ for \emnist. The x-axis represents frequency intervals and the y-axis represents the number of frequencies in that interval. (d) The performance of TLSTM3+\timetovec\ on \culike\ in terms of \recallten\ with and without the linear term. 
   }
\end{figure}

We argued in Section~\ref{sec:time2vec} that a representation of time needs to capture both periodic and non-periodic patterns. The sine functions in \timetovec\ can be used to capture periodic patterns and the linear term can be used to capture non-periodic ones. In Section~\ref{comparison-results}, we showed that capturing both periodic and non-periodic patterns (\ie LSTM+\timetovec) gives better performance than only capturing non-periodic patterns (\ie LSTM+T). A question that remains to be answered is whether capturing both periodic and non-periodic patterns gives better performance than only capturing periodic patterns. To answer this question, we first repeated the experiment for \emnist\ when the linear term is removed from \timetovec\ but observed that the results were not affected substantially. This means capturing periodic patterns may suffice for \emnist. Then we conducted a similar experiment for TLSTM3 on \culike\ and obtained the results in Fig.~\ref{fig:activation-fix-freq}(d). From these results, we can see that capturing both periodic and non-periodic patterns makes a model perform better than only capturing periodic patterns. This experiment can also be viewed as an ablation study indicating why a linear term is required in \timetovec.

\section{Conclusion \& Future Work}
In a variety of tasks for synchronous or asynchronous event predictions, time is an important feature with specific characteristics (progression, periodicity, scale, etc.).   
In this work, we presented an approach that automatically learns features of time that represent these characteristics.
In particular, we developed \timetovec, a vector representation for time, using \sine\ and linear activations and showed the effectiveness of this representation across several datasets and several tasks. In the majority of our experiments, \timetovec\ improved our results, while the remaining results were not hindered by its application. 
While \sine\ functions have been argued to complicate the optimization \cite{lapedes1987nonlinear,taming2017}, we did not experience such a complication except for the experiment in Subsection~\ref{syn-exp-section} when using only a few \sine\ functions. We hypothesize that the main reasons include combining \sine\ functions with a powerful model (\eg, LSTM) and using many \sine\ functions which reduces the distance to the goal (see, \eg, \cite{neyshabur2018role}). We leave a deeper theoretical analysis of this hypothesis and the development of better optimizers as future work. 
\bibliography{asynchronous}
\bibliographystyle{named}

\appendix

\begin{figure}[t]
   \centering
   \includegraphics[width=0.38\textwidth]{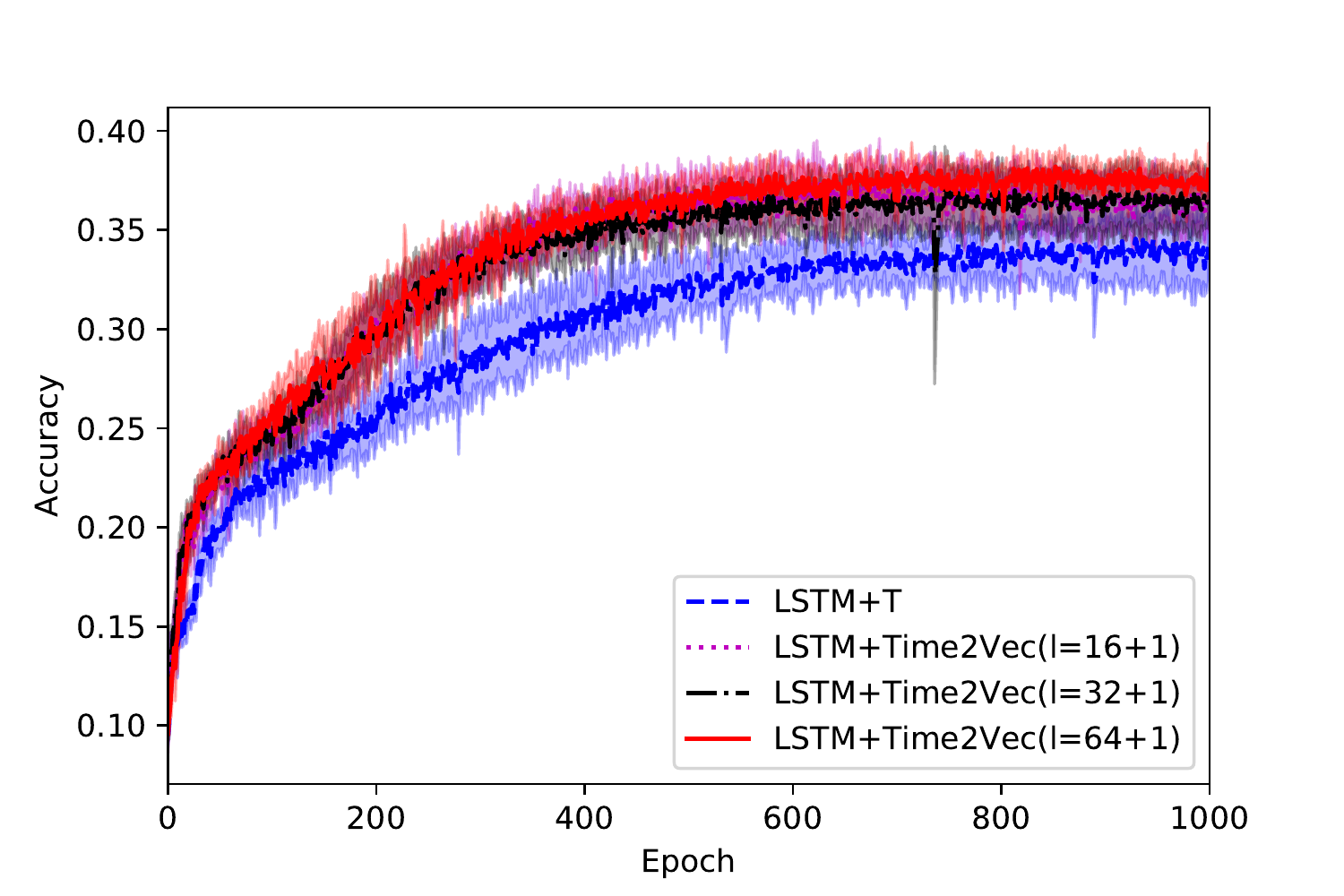}
   \caption{%
   \label{fig:ntdigits} %
   Comparing LSTM+T and LSTM+\timetovec\ on \emnist.}
\end{figure}

\begin{figure}[bp!]
   \centering
   \subfloat[\emnist]{%
   \includegraphics[width=0.38\textwidth]{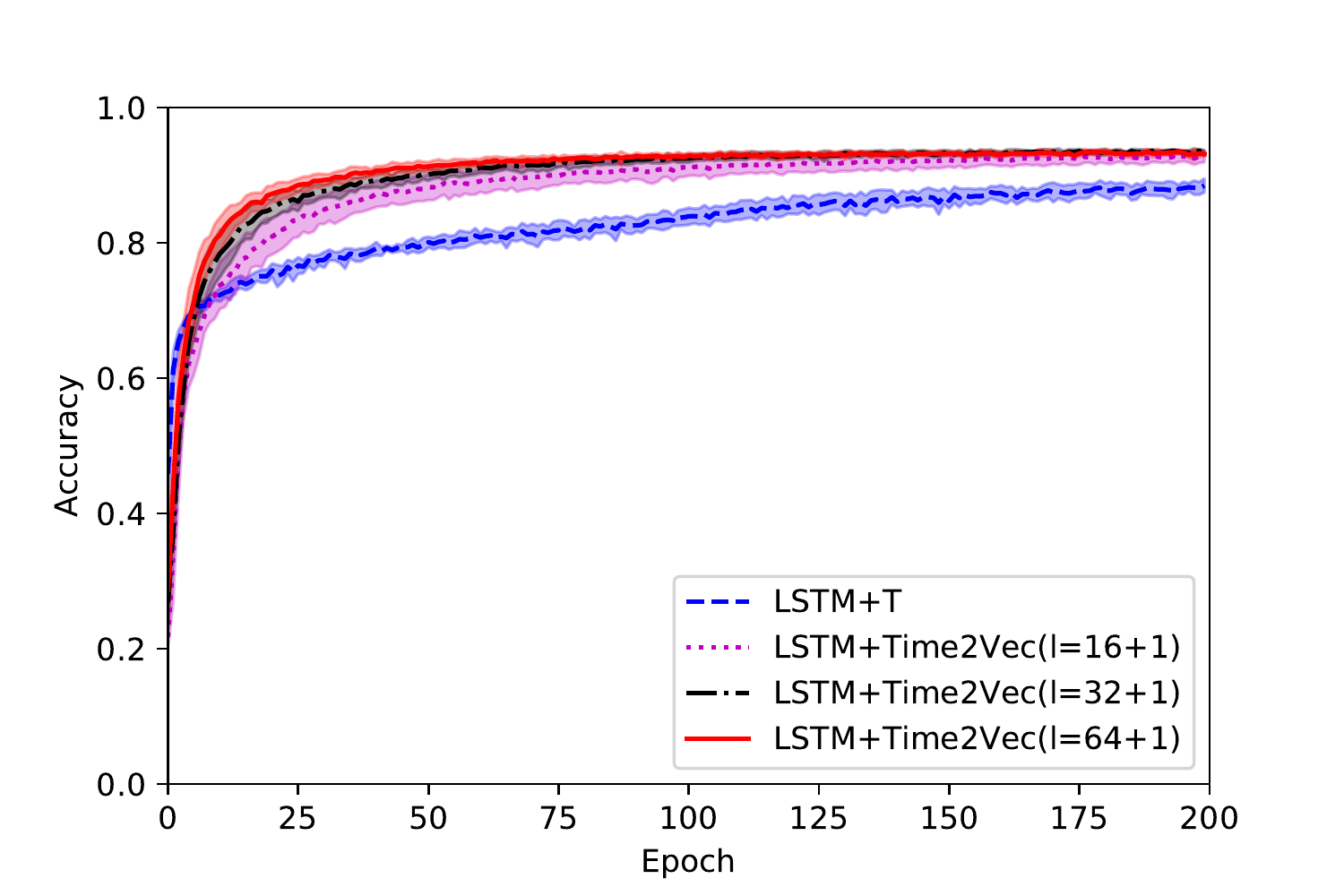}}
~~~~\hspace*{0cm}
   \subfloat[Raw \ntdigits]{%
   \includegraphics[width=0.38\textwidth]{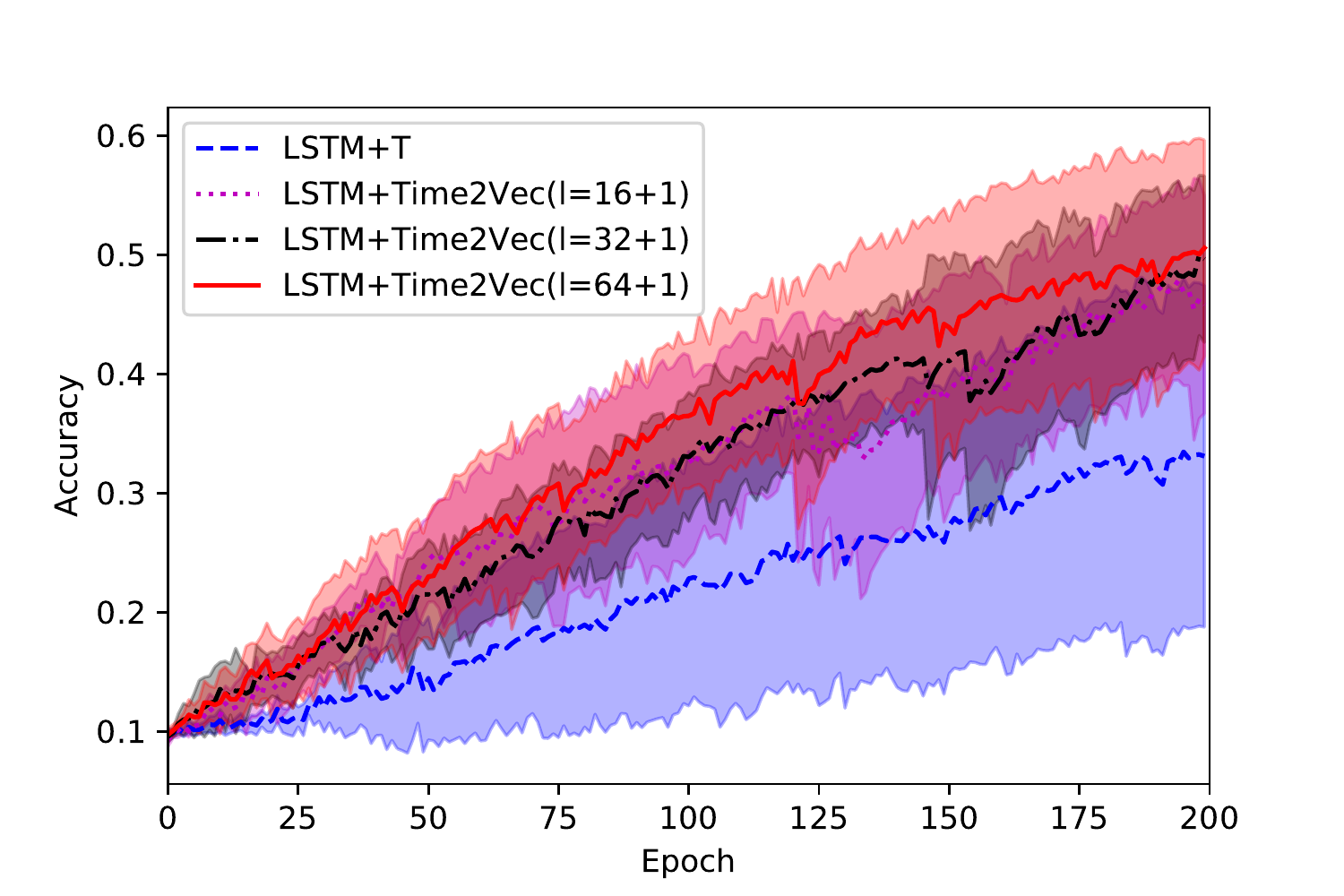}} 
   \caption{%
   \label{fig:mnist-ntdigits} %
   Comparing LSTM+T and LSTM+\timetovec\ on \emnist\ and raw \ntdigits.}
\end{figure}

\begin{figure*}[t]
  \centering
\includegraphics[width=0.8\textwidth]{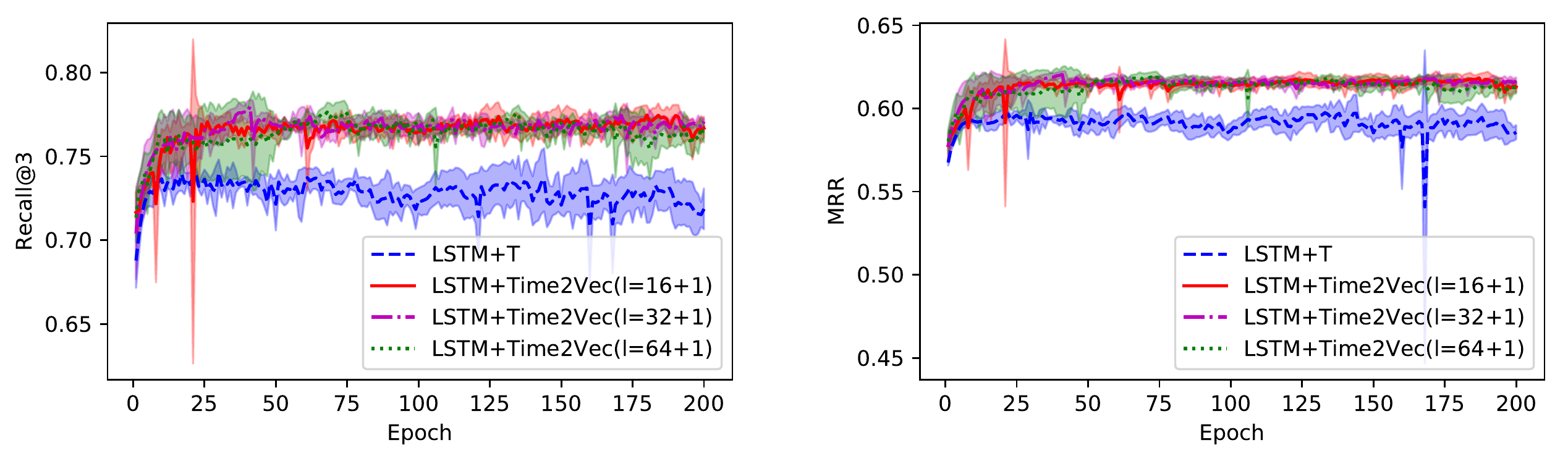}
  \caption{Comparing LSTM+T and LSTM+\timetovec\ on SOF.}
  \label{fig:SOF}
\end{figure*}

\begin{figure*}[t]
  \centering
\includegraphics[width=0.8\textwidth]{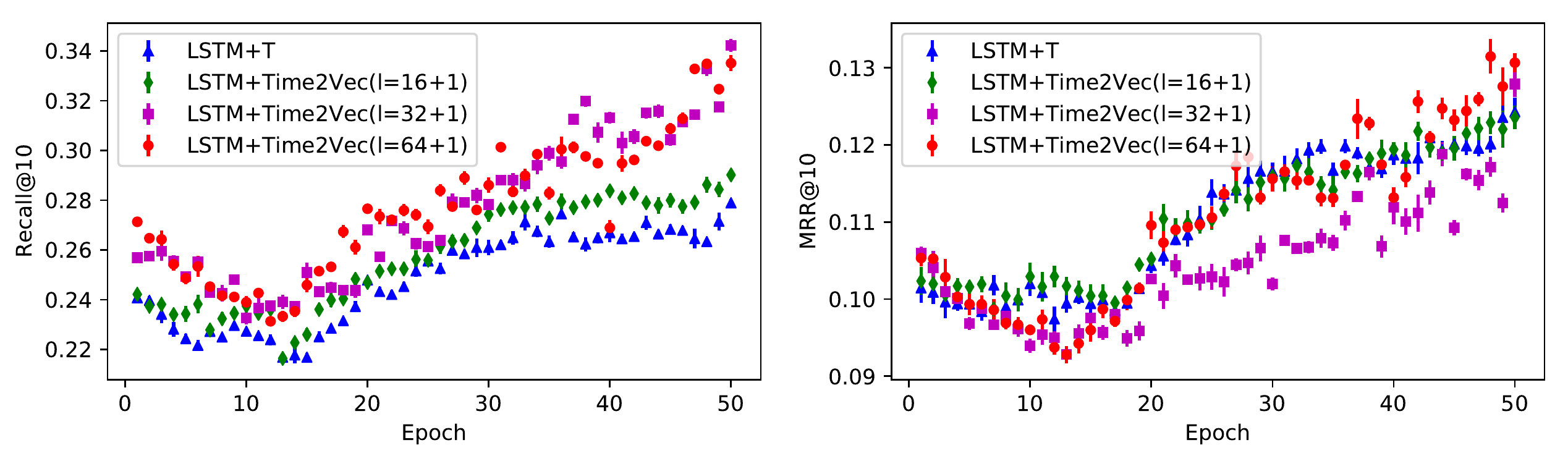}
  \caption{Comparing LSTM+T and LSTM+\timetovec\ on \lastfm.}
  \label{tlstm-t-lastfm-fig}
\end{figure*}

\begin{figure*}[t]
  \centering
\includegraphics[width=0.8\textwidth]{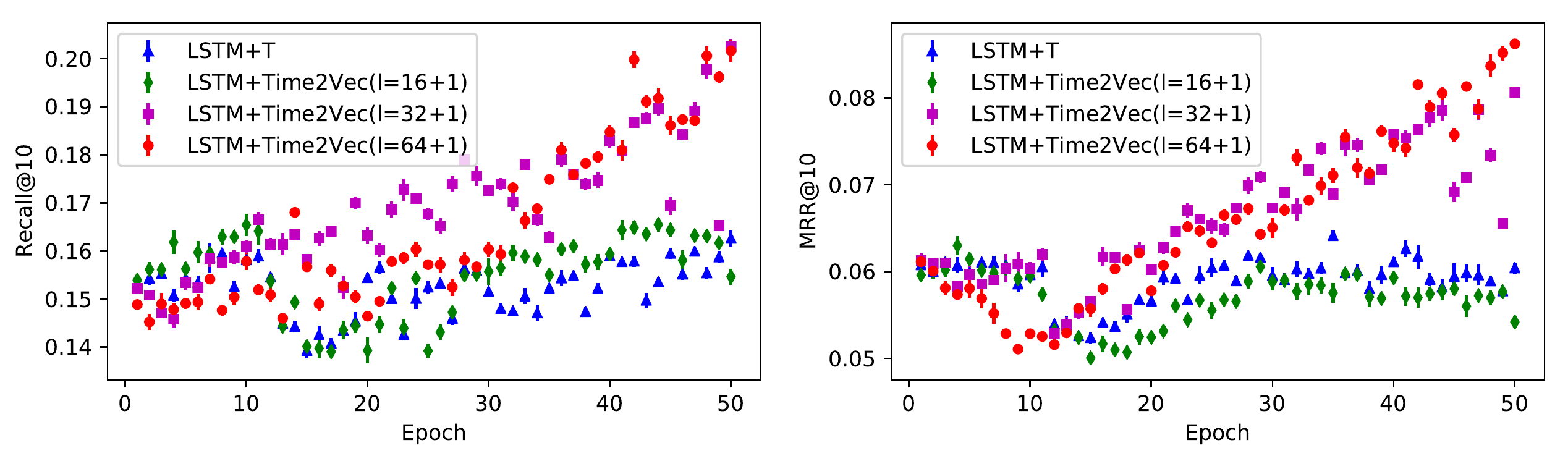}
  \caption{Comparing LSTM+T and LSTM+\timetovec\ on \culike.}
  \label{tlstm-t-culike-fig}
\end{figure*}

\section{Implementation Detail} \label{appendix:implementation}
For the experiments on \emnist, \ntdigits\ and SOF, we implemented\footnote{Code and datasets are available at: \url{https://github.com/borealisai/Time2Vec}} our model in PyTorch \cite{paszke2017automatic}. We used Adam optimizer \cite{kingma2014adam} with a learning rate of $0.001$. For \emnist\ and SOF, we fixed the hidden size of the LSTM to $128$. For \ntdigits, due to its smaller train set, we fixed the hidden size to $64$. We allowed each model $200$ epochs.
We used a batch size of $512$ for \emnist\ and $128$ for \ntdigits\ and SOF. 
For the experiments on \lastfm\ and \culike, we used the code released by \citet{zhu2017next}\footnote{\url{https://github.com/DarryO/time_lstm}} without any modifications, except replacing $\tau$ with $\ttov(\tau)$. The only other thing we changed in their code was to change the \emph{SAMPLE\_TIME} variable from $3$ to $20$. \emph{SAMPLE\_TIME} controls the number of times we do sampling to compute \recallten\ and \mrrten. We experienced a high variation when sampling only $3$ times so we increased the number of times we sample to $20$ to make the results more robust. For both \lastfm\ and \culike, Adagrad optimizer is used with a learning rate of $0.01$, vocabulary size is $5000$, and the maximum length of the sequence is $200$. 
For \lastfm, the hidden size of the LSTM is $128$ and for \culike, it is $256$. For all except the synthesized dataset, we shifted the event times such that the first event of each sequence starts at time $0$.

For the fairness of the experiments, we made sure the competing models for all our experiments have an (almost) equal number of parameters. For instance, since adding \timetovec\ as an input to the LSTM increases the number of model parameters compared to just adding time as a feature, we reduced the hidden size of the LSTM for this model to ensure the number of model parameters stays (almost) the same. For the experiments involving \timetovec, unless stated otherwise, we tried vectors with $16$, $32$ and $64$ \sine\ functions (and one linear term). We reported the vector length offering the best performance in the main text. The results for other vector lengths can be found in Appendix~\ref{appendix:more-results}. For the synthetic dataset, we use Adam optimizer with a learning rate of $0.001$ without any regularization. The length of the \timetovec\ vector is $32$. 

\section{More Results} \label{appendix:more-results}
We ran experiments on other versions of the \ntdigits\ dataset as well. Following~\cite{anumula2018feature}, we converted the raw event data to \emph{event-binned} features by virtue of aggregating active channels through a period of time in which a pre-defined number of events occur. The outcome of binning is thus consecutive frames each with multiple but a fixed number of active channels. In our experiments, we used event-binning with 100 events per frame. For this variant of the dataset, we compared LSTM+T and LSTM+\timetovec\ similar to the experiments in Section~\ref{comparison-results}. The obtained results were on-par. Then, similar to \emnist, we only fed as input the times at which events occurred (\ie we removed the channels from the input). We allowed the models $1000$ epochs to make sure they converge. The obtained results are presented in Fig.~\ref{fig:ntdigits}. It can be viewed that \timetovec\ provides an effective representation for time and LSTM+\timetovec\ outperforms LSTM+T on this dataset.

\begin{figure*}[t]
  \centering
\includegraphics[width=0.8\textwidth]{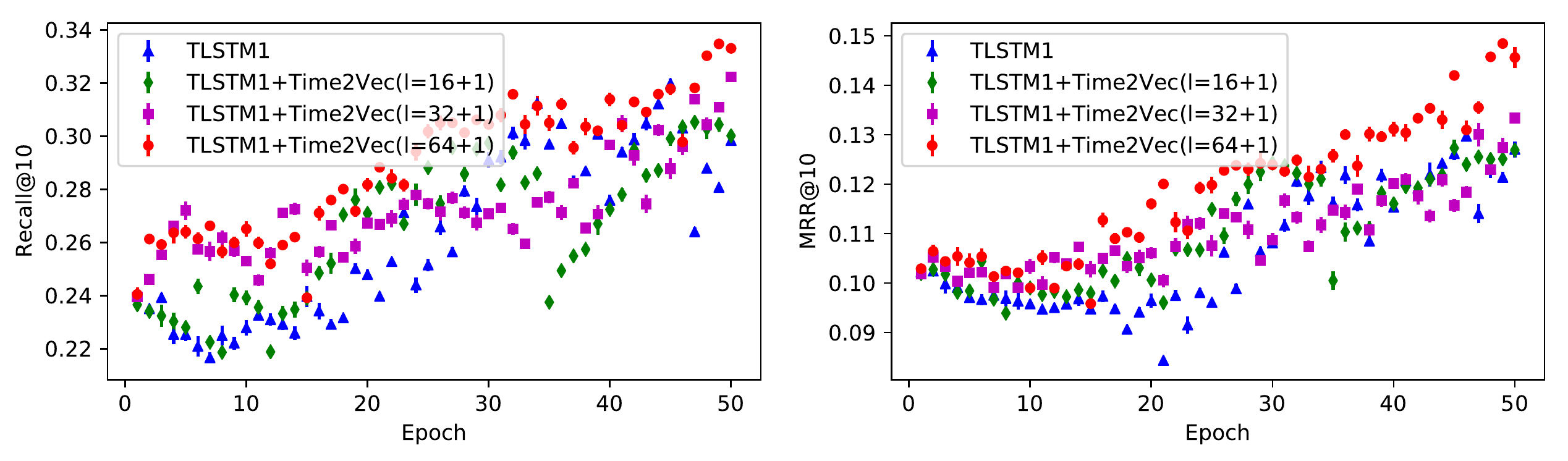}
  \caption{TLSTM1's performance on \lastfm\ with and without \timetovec.}
  \label{tlstm1-lastfm-fig}
\end{figure*}

\begin{figure*}[t]
  \centering
\includegraphics[width=0.8\textwidth]{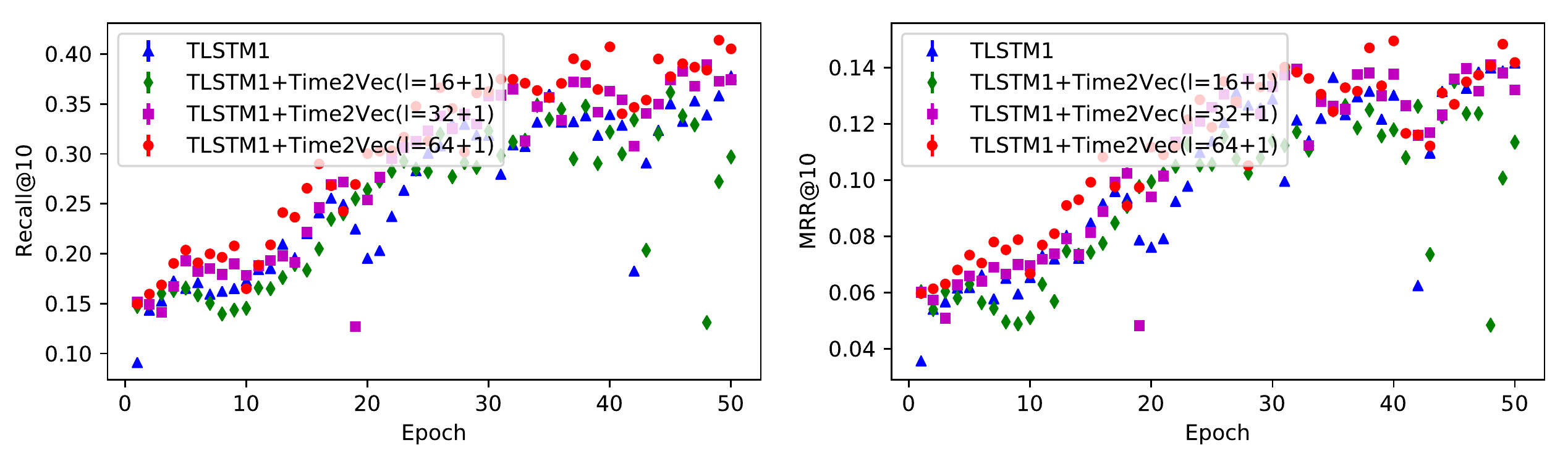}
  \caption{TLSTM1's performance on \culike\   with and without \timetovec.}
  \label{tlstm1-citeulike-fig}
\end{figure*}

\begin{figure*}[tp!]
  \centering
\includegraphics[width=0.8\textwidth]{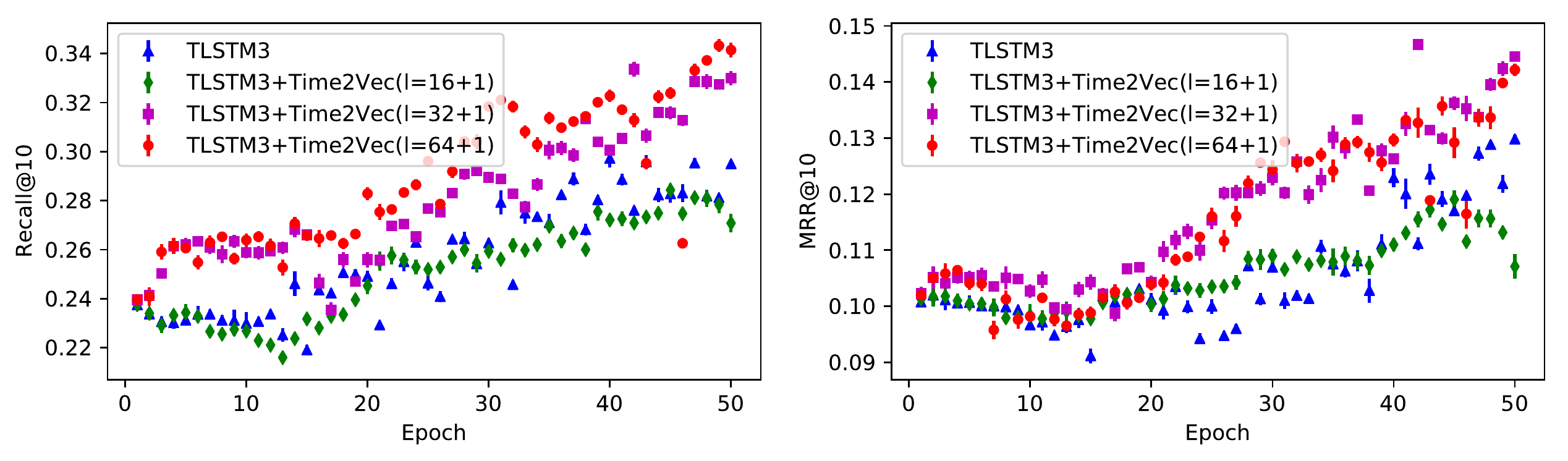}
  \caption{TLSTM3's performance on \lastfm\   with and without \timetovec.}
  \label{tlstm3-lastfm-fig}
\end{figure*}

\begin{figure*}[tp!]
  \centering
\includegraphics[width=0.8\textwidth]{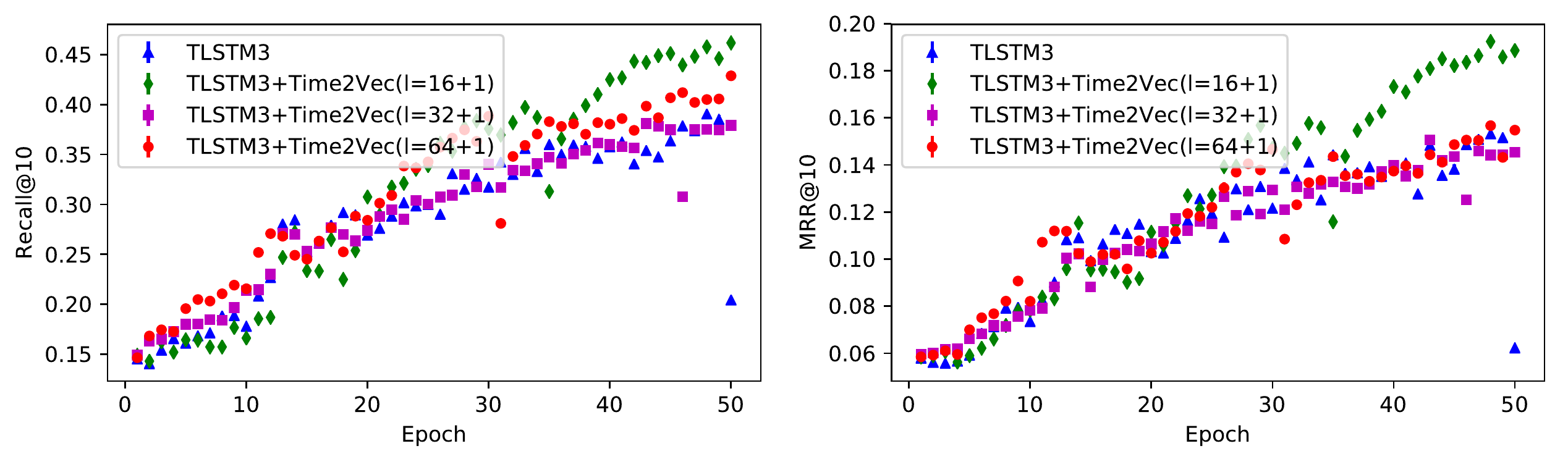}
  \caption{TLSTM3's performance on \culike\   with and without \timetovec.}
  \label{tlstm3-citeulike-fig}
\end{figure*}

In the main text, for the experiments involving \timetovec, we tested \timetovec\ vectors with $16$, $32$ and $64$ sine functions and reported the best one for the clarity of the diagrams. Here, we show the results for all frequencies. Figures~\ref{fig:mnist-ntdigits},~\ref{fig:SOF},~\ref{tlstm-t-lastfm-fig},~and~\ref{tlstm-t-culike-fig} compare LSTM+T and LSTM+\timetovec\ for our datasets. Figures~\ref{tlstm1-lastfm-fig},~and~\ref{tlstm1-citeulike-fig} compare TLSTM1 with TLSTM1+\timetovec\ on \lastfm\ and \culike. Figures~\ref{tlstm3-lastfm-fig},~and~\ref{tlstm3-citeulike-fig} compare TLSTM3 with TLSTM1+\timetovec\ on \lastfm\ and \culike. In most cases, \timetovec\ with 64 sine functions outperforms (or gives on-par results with) the cases with 32 or 16 sine functions. An exception is TLSTM3 where 16 sine functions works best. We believe that is because TLSTM3 has two time gates and adding, e.g., 64 temporal components (corresponding to the sine functions) to each gate makes it overfit to the temporal signals. 

\section{LSTM Architectures} \label{appendix:lstm}
The original LSTM model can be neatly defined with the following equations:
\begin{align}
&\vctr{i}_j= \sigma\left(\mtrx{W}_i
\vctr{x}_j+\mtrx{U}_i\vctr{h}_{j-1}+\vctr{b}_i\right) \label{lstm-eq-i}\\
&\vctr{f}_j= \sigma\left(\mtrx{W}_f\vctr{x}_j+\mtrx{U}_f\vctr{h}_{j-1}+\vctr{b}_f\right)\label{lstm-eq-f}\\
&\overline{\vctr{c}}_j = \tanhyp\left(\mtrx{W}_c \vctr{x}_j+\mtrx{U}_c \vctr{h}_{j-1}+\vctr{b}_c\right) \label{lstm-eq-cbar}\\
&\vctr{c}_j = \vctr{f}_t\odot\vctr{c}_{j-1} + \vctr{i}_j\odot \overline{\vctr{c_j}} \label{lstm-eq-c}\\
&\vctr{o}_j= \sigma\left(\mtrx{W}_o \vctr{x}_j+\mtrx{U}_o \vctr{h}_{j-1}+\vctr{b}_o\right)\label{lstm-eq-o}\\
&\vctr{h}_j= \vctr{o}_j\odot\tanhyp\left(\vctr{c}_j\right)\label{lstm-eq-h}
\end{align}
Here $\vctr{i}_t$, $\vctr{f}_t$, and $\vctr{o}_t$ represent the input, forget and output gates respectively, while $\vctr{c}_t$ is the memory cell and $\vctr{h}_t$ is the hidden state. $\sigma$ and $\tanhyp$ represent the \sigmoid\ and hyperbolic tangent activation functions respectively. We refer to $\vctr{x}_j$ as the $j^{th}$ event. 

\textbf{Peepholes:} \citet{gers2000recurrent} introduced a variant of the LSTM architecture where the input, forget, and output gates peek into the memory cell. In this variant, $\vctr{w}_{pi} \odot \vctr{c}_{j-1}$,  $\vctr{w}_{pf} \odot \vctr{c}_{j-1}$, and $\vctr{w}_{po} \odot \vctr{c}_{j}$ are added to the linear parts of Eq.~(\ref{lstm-eq-i}),~(\ref{lstm-eq-f}),~and~(\ref{lstm-eq-o}) respectively, where $\vctr{w}_{pi}$, $\vctr{w}_{pf}$, and $\vctr{w}_{po}$ are learnable parameters.

\textbf{LSTM+T:} Let $\vctr{\tau}_j$ represent the time features for the $j^{th}$ event in the input and let $\vctr{x}'_j = [\vctr{x}_j;\vctr{\tau}_j]$. Then LSTM+T uses the exact same equations as the standard LSTM (denoted above) except that $\vctr{x}_j$ is replaced with $\vctr{x}'_j$. 

\textbf{TimeLSTM:} 
We explain TLSTM1 and TLSTM3 which have been used in our experiments. For clarity of writing, we do not include the peephole terms in the equations but they are used in the experiments.
In TLSTM1, a new time gate is introduced as in Eq.~\eqref{tlstm1-time-gate-eq} and Eq.~\eqref{lstm-eq-c}~and~\eqref{lstm-eq-o} are updated to Eq.~\eqref{tlstm1-eq-c}~and~\eqref{tlstm1-eq-o} respectively: 
\begin{align}
&\vctr{t}_j = \sigma \left(\mtrx{W}_t \vctr{x}_j + \sigma \left(\vctr{u}_t \tau_j \right)+\vctr{b}_t \right)  \label{tlstm1-time-gate-eq}\\
&\vctr{c}_j = \vctr{f}_j\odot\vctr{c}_{j-1} + \vctr{i}_j\odot\vctr{t}_j\odot \overline{\vctr{c_j}}\label{tlstm1-eq-c}\\
&\vctr{o}_j=\sigma\left(\mtrx{W}_o \vctr{x}_j+\vctr{v}_t\tau_j +\mtrx{U}_o \vctr{h}_{j-1}+\vctr{b}_o\right)\label{tlstm1-eq-o}
\end{align}
$\vctr{t}_j$ controls the influence of the current input on the prediction and makes the required information from timing history get stored on the cell state.
TLSTM3 uses two time gates: 
\begin{align} 
&\vctr{t1}_j = \sigma\left(\mtrx{W}_{t1}
\vctr{x}_j+\sigma\left(\vctr{u}_{t1}\tau_j\right)+\vctr{b}_{t1}\right) \label{tlstm3-t1-eq}\\
&\vctr{t2}_j = \sigma\left(\mtrx{W}_{t2}
\vctr{x}_j+\sigma\left(\vctr{u}_{t2}\tau_j\right)+\vctr{b}_{t2}\right)\label{tlstm3-t2-eq}
\end{align}
where the elements of $\mtrx{W}_{t1}$ are constrained to be non-positive. $\vctr{t1}$ is used for controlling the influence of the last consumed item and $\vctr{t2}$  stores the $\tau$s thus enabling modeling long range dependencies.
TLSTM3 couples the input and forget gates following \citet{greff2017lstm} along with the $\vctr{t1}$ and $\vctr{t2}$
gates and replaces Eq.~(\ref{lstm-eq-c})~to~(\ref{lstm-eq-h}) with the following:
\begin{align}
&\widetilde{\vctr{c}}_j = (1-\vctr{i}_j\odot\vctr{t1}_j)\odot\vctr{c}_{j-1} + \vctr{i}_j\odot\vctr{t1}_j\odot \overline{\vctr{c}}_j\label{tlstm3-eq-c-tilde}\\
&\vctr{c}_j =(1-\vctr{i}_j)\odot\vctr{c}_{j-1} + \vctr{i}_j\odot\vctr{t2}_j\odot \overline{\vctr{c}}_j\label{tlstm3-eq-c}\\
&\vctr{o}_j=\sigma\left(\mtrx{W}_o \vctr{x}_j+\vctr{v}_t\tau_j +\mtrx{U}_o \vctr{h}_{j-1}+\vctr{b}_o\right)\label{tlstm3-eq-o}\\
&\vctr{h}_j = \vctr{o}_j\odot\tanhyp\left(\widetilde{\vctr{c}}_j\right)\label{tlstm3-eq-h}
\end{align}
\citet{zhu2017next} use $\tau_j=\DeltaT_j$ in their experiments, where $\DeltaT_j$ is the duration between the current and the last event.

\textbf{TimeLSTM+\timetovec:} 
To replace time in TLSTM1 with \timetovec, we modify Eq.~(\ref{tlstm1-time-gate-eq})~and~(\ref{tlstm1-eq-o}) as follows:
\begin{align}
&\vctr{t}_j = \sigma\left(\mtrx{W}_t
\vctr{x}_j+\sigma\left(\mtrx{U}_t\ttov(\tau)\right)+\vctr{b}_t\right)\label{tlstm1-time-gate-modified-eq}\\
&\vctr{o}_j=\sigma(\mtrx{W}_o \vctr{x}_j+\mtrx{V}_t\ttov(\tau) +\mtrx{U}_o \vctr{h}_{j-1}+\vctr{b}_o)
\end{align}
i.e., $\tau$ is replaced with $\ttov(\tau)$, $\vctr{u}_t$ is replaced with $\mtrx{U}_t$, and $\vctr{v}_t$ is replaced with $\mtrx{V}_t$. 
Similarly, for TLSTM3 we modify Eq.~(\ref{tlstm3-t1-eq}),~(\ref{tlstm3-t2-eq}) and~(\ref{tlstm3-eq-o}) as follows:
\begin{align}
    \vctr{t1}_j = \sigma\left(\mtrx{W}_{t1}
\vctr{x}_j+\sigma\left(\mtrx{U}_{t1}\ttov(\tau)\right)+\vctr{b}_{t1}\right)\label{tlstm3-time-gate-1-modified-eq}\\
    \vctr{t2}_j = \sigma\left(\mtrx{W}_{t2}
\vctr{x}_j+\sigma\left(\mtrx{U}_{t2}\ttov(\tau)\right)+\vctr{b}_{t2}\right)\label{tlstm3-time-gate-2-modified-eq}\\
    \vctr{o}_j=\sigma(\mtrx{W}_o \vctr{x}_j+\mtrx{V}_t\ttov(\tau)
+\mtrx{U}_o \vctr{h}_{j-1}+\vctr{b}_o)\label{tlstm3-t2v-o}
\end{align}

\section{Proofs}
\setcounter{proposition}{0}
\label{appendix:proofs}
\begin{proposition}
\timetovec\ is invariant to time rescaling.
\end{proposition}

\begin{proof}
Consider the following \timetovec\ representation $\mathcal{M}_1$:
\begin{equation}
    \label{eq:t2vec_def}
  \ttov(\tau)[i]=\begin{cases}
    \omega_i \tau + \varphi_i, & \text{if~~$i=0$}. \\
    \sin{(\omega_i \tau + \varphi_i)}, & \text{if~~$1\leq i \leq k$}.
  \end{cases}
\end{equation} 
Replacing $\tau$ with $\alpha \cdot \tau$ (for $\alpha > 0$), the \timetovec\ representation updates as follows:
\begin{equation}
    \label{eq:t2vec_def}
  \ttov(\alpha\cdot\tau)[i]=\begin{cases}
    \omega_i (\alpha\cdot\tau) + \varphi_i, & \text{if~~$i=0$}. \\
    \sin{(\omega_i (\alpha\cdot\tau) + \varphi_i)}, & \text{if~~$1\leq i \leq k$}.
  \end{cases}
\end{equation}
Consider another \timetovec\ representation $\mathcal{M}_2$ with frequencies $\omega'_i=\frac{\omega_i}{\alpha}$. Then $\mathcal{M}_2$ behaves in the same way as $\mathcal{M}_1$. This proves that \timetovec\ is invariant to time rescaling.
\end{proof}

\end{document}